\pdfoutput=1

\documentclass[11pt]{article}

\usepackage[final]{acl}

\usepackage{times}
\usepackage{latexsym}

\usepackage[T1]{fontenc}

\usepackage[utf8]{inputenc}

\usepackage{microtype}

\usepackage{inconsolata}

\usepackage{graphicx}
\usepackage{amsmath}
\usepackage{amsthm}
\usepackage{amsfonts}
\usepackage{amssymb}
\usepackage{tikz}
\usepackage{cleveref}
\usepackage{booktabs}
\usepackage{bbm}
\usepackage{mathtools}
\usepackage{multirow}
\usepackage{color,soul}
\usepackage{float} 
\usepackage{makecell}
\usepackage{array}
\newcolumntype{P}[1]{>{\raggedright\arraybackslash}p{#1}}

\usepackage{titlesec}
\titlespacing*{\paragraph}{0ex}{0.5ex}{1ex}

\usepackage{enumitem}
\setlist[enumerate,itemize]{topsep=0pt,itemsep=0pt,leftmargin=18pt}

\usepackage{soul}

\usepackage{subcaption}
\usepackage{adjustbox}
\usepackage{examples-slim}

\theoremstyle{definition}

\DeclareMathSymbol{\shortminus}{\mathbin}{AMSa}{"39}
\usepackage{makecell} 

\crefformat{section}{\S#2#1#3}
\crefformat{subsection}{\S#2#1#3}
\crefformat{subsubsection}{\S#2#1#3}
\crefformat{paragraph}{\P#2#1#3}
\crefformat{subparagraph}{\P#2#1#3}
\crefmultiformat{section}{\S#2#1#3}{ and~\S#2#1#3}{, \S#2#1#3}{, and~\S#2#1#3}
\crefmultiformat{subsection}{\S#2#1#3}{ and~\S#2#1#3}{, \S#2#1#3}{, and~\S#2#1#3}
\crefmultiformat{subsubsection}{\S#2#1#3}{ and~\S#2#1#3}{, \S#2#1#3}{, and~\S#2#1#3}
\crefmultiformat{paragraph}{\P\P#2#1#3}{ and~#2#1#3}{, #2#1#3}{, and~#2#1#3}
\crefmultiformat{subparagraph}{\P\P#2#1#3}{ and~#2#1#3}{, #2#1#3}{, and~#2#1#3}
\crefrangeformat{section}{\mbox{\S\S#3#1#4--#5#2#6}}
\crefrangeformat{subsection}{\mbox{\S\S#3#1#4--#5#2#6}}
\crefrangeformat{subsubsection}{\mbox{\S\S#3#1#4--#5#2#6}}
\crefrangeformat{paragraph}{\mbox{\P\P#3#1#4--#5#2#6}}
\crefrangeformat{subparagraph}{\mbox{\P\P#3#1#4--#5#2#6}}
\crefname{part}{Part}{Parts}
\Crefname{part}{Part}{Parts}
\crefname{chapter}{Ch.}{Ch.}
\Crefname{chapter}{Ch.}{Ch.}
\crefname{footnote}{Fn.}{Fn.}
\Crefname{footnote}{Fn.}{Fn.}
\crefname{figure}{Figure}{Figures}
\crefname{table}{Table}{Tables}
\crefname{subfigure}{Figure}{Figures}
\Crefname{subfigure}{Figure}{Figures}
\crefname{appsec}{Appendix}{Appendices}
\Crefname{appsec}{Appendix}{Appendices}
\crefname{algocf}{Algorithm}{Algorithms}
\Crefname{algocf}{Algorithm}{Algorithms}
\crefname{xnumi}{ex.}{exs.}
\Crefname{xnumi}{Ex.}{Exs.}
\crefname{xnumii}{ex.}{exs.}
\Crefname{xnumii}{Ex.}{Exs.}

\title{
Causal Interventions Reveal Shared Structure\\[.2ex]Across English Filler--Gap Constructions}

\author{Sasha Boguraev$^1$ \quad Christopher Potts$^2$ \quad Kyle Mahowald$^1$ \\
        $^1$The University of Texas at Austin \quad $^2$Stanford University\\
        \texttt{\{sasha.boguraev,kyle\}@utexas.edu} \qquad \texttt{cgpotts@stanford.edu}
        }

\begin{document}
\maketitle
\begin{abstract}
Language Models (LMs) have emerged as powerful sources of evidence for linguists seeking to develop theories of syntax. In this paper, we argue that causal interpretability methods, applied to LMs, can greatly enhance the value of such evidence by helping us characterize the abstract mechanisms that LMs learn to use. Our empirical focus is a set of English filler--gap dependency constructions (e.g., questions, relative clauses). Linguistic theories largely agree that these constructions share many properties. Using experiments based in Distributed Interchange Interventions, we show that LMs converge on similar abstract analyses of these constructions. These analyses also reveal previously overlooked factors -- relating to frequency, filler type, and surrounding context -- that could motivate changes to standard linguistic theory. Overall, these results suggest that mechanistic, internal analyses of LMs can push linguistic theory forward. 
\newline
\newline
\hspace{.5em}\includegraphics[width=1.25em,height=1.25em]{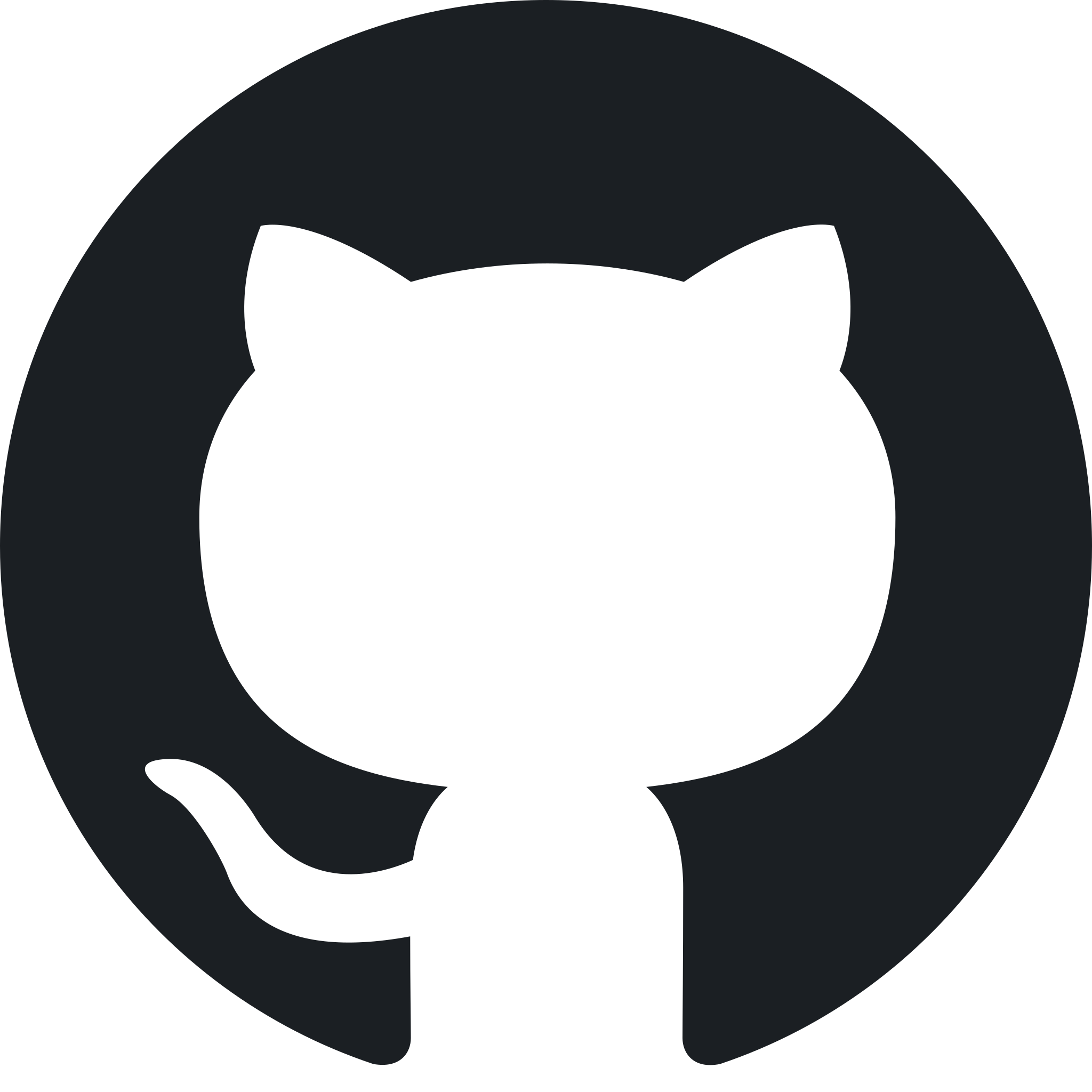}\hspace{.75em}\parbox{\dimexpr\linewidth-2\fboxsep-2\fboxrule}{\url{https://github.com/SashaBoguraev/causal-filler-gap}}
\end{abstract}

\section{Introduction}

Language models can generate and process utterances typically thought to require rich linguistic grammatical structure \citep{futrell2019syntactic,wilcox-etal-2018-rnn,manning2020emergent,hu2020systematic}, including much-studied syntactic constructions like long-distance filler--gap constructions \citep{wilcox2024using}. These results have been taken to challenge claims that these phenomena can be learned only with strong innate priors \citep{piantadosi2023modern,futrell2025linguistics}.

\begin{figure}[t]
    \centering
    \includegraphics[width=.98\linewidth]{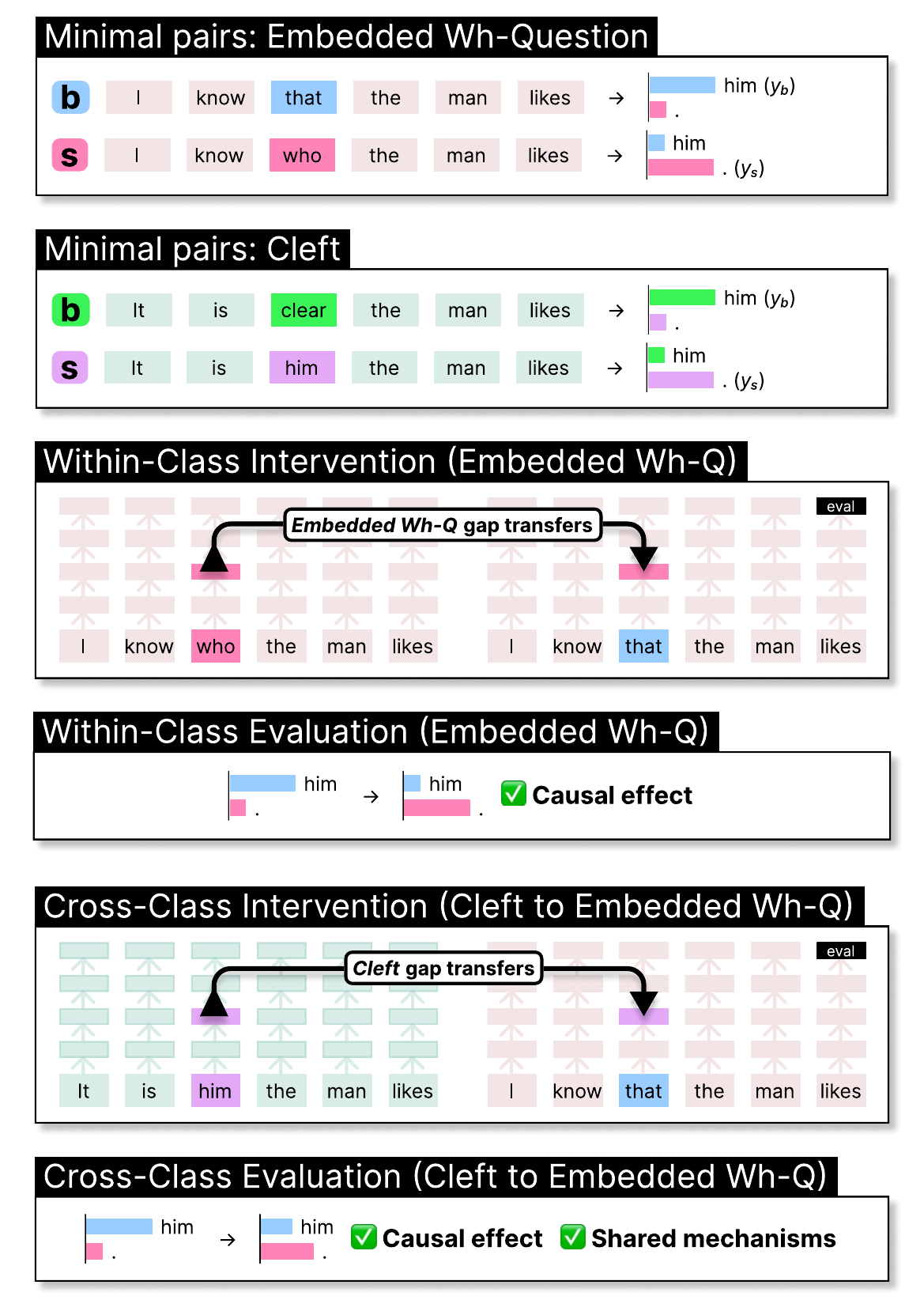}
    \caption{\textbf{Causal intervention overview.} Here, we illustrate our methodology when we intervene within a class, transferring an embedded wh- filler--gap structure into a corresponding minimal pair that didn't previously have one. We then show intervening across classes, inserting a wh- filler--gap into a gap-less cleft sentence.}
    \label{fig:schematic}
\end{figure}

Despite the strong performance, questions remain as to whether models acquire syntax in ways that are posited by linguists to be human-like (e.g., acquiring rich grammatical abstraction and syntactic structure).
Causal interpretability methods now make it possible to characterize the abstract mechanisms underlying neural networks \citep{vig2020causal,finlayson-etal-2021-causal,Geiger:Lu-etal:2021,meng2022locating,geiger2023causal,wang2023interpretability}.
These methods have revealed non-trivial linguistic syntactic structure is learned by models \citep{lakretz2019emergence,finlayson-etal-2021-causal,mueller-etal-2022-causal,lasri-etal-2022-probing,arora-etal-2024-causalgym}.
But a key hypothesis in the history of linguistics is that seemingly different linguistic constructions can share underlying structure.
For instance, compare ``I wonder what the lion ate.''\ to ``It was the gazelle that the lion ate.'' 
The former is an embedded wh- clause and the latter is a cleft construction.
These are distinct constructions but share something in common: both have a long-distance dependency with an extracted element, often specified with a linguistic trace: ``I wonder what$_t$ the lion ate \_\_\_$_t$.'' and ``It was the gazelle$_t$ that the lion ate \_\_\_$_t$.''
Thus, many linguistic theories predict common processing characteristics between these sentences \citep{fodor1989empty}. 
On the other hand, there is also reason to expect wh- sentences to be quite different from clefts since both wh- elements  and clefts have idiosyncratic properties \citep{ross1967constraints,Culicover99NUTS,Ginzburg00}.

To tackle these questions, we take advantage of advances in large open source models as well as in mechanistic interpretability, specifically the Causal Abstraction framework \citep{geiger2023causal} and Distributed Alignment Search (DAS; \citealt{geiger2024finding}). Our resulting methodology gives us direct access to the abstract causal mechanisms learned by these models. By accessing these causal mechanisms, we can take a filler-gap mechanism learned on Construction A (e.g., wh- sentences), transfer it to Construction B (e.g., clefts), and see if we get predictable filler--gap behavior (see \Cref{fig:schematic}). If we do, this would be strong evidence of underlying shared structure learned by the model.

Importantly, this method gives us a gradient measure of transfer. As such, we explore whether more similar constructions transfer more readily to each other; whether some constructions in general tend to serve as sources of transfer; whether mechanisms transfer across clauses; and whether transfer is greater when lexical items are shared across constructions \citep[an effect predicted by the ``lexical boost'' in syntactic priming, whereby syntactic structures are primed more strongly when there is lexical overlap;][]{pickering1998representation}.

Ultimately, we find strong generalization in LMs across a range of filler--gap constructions, with effects observed at all positions within constructions. 
We observe lexical boost: effects are stronger when lexical items match (e.g., the same animacy).
We find greater transfer between constructions that share linguistically relevant features (e.g., the nature of the filler or whether fronting alters the utterance’s information structure).
Moreover, we identify \textit{source} constructions whose underlying mechanisms generalize broadly, as well as \textit{sink} constructions that consistently benefit from such transferred mechanisms.
Finally, we provide evidence that such generalization does \textit{not} seem to extend across clausal boundaries.

We claim these experiments make good on the promise that studying LMs can help us better understand linguistic structure and language learning in general by not just serving as proxies for data-driven learners, but by helping us develop linguistically interesting hypotheses \citep{potts2023characterizing,futrell2025linguistics}.

\begin{table*}[t]
\footnotesize
\centering
\setlength{\tabcolsep}{2pt}
\renewcommand{\arraystretch}{1.1}
\resizebox{\textwidth}{!}{%
\begin{tabular}{@{}P{2.6cm}lllllll|cccc@{}}
\toprule
\textbf{Construction} & \textbf{Prefix} & \textbf{Filler} & \textbf{NC} & \textbf{Article} & \textbf{NP} & \textbf{Verb} & \textbf{Label} &
\textbf{Filler} & \textbf{Inverted} &
\textbf{Embed} &
\textbf{Front} \\
\midrule
Emb.~Wh-Q (\textit{Know})   & I know   & who/that &    & the & man & liked & ./him   & Wh   & No  & VP  & No  \\
\mbox{Emb. Wh- (\textit{Wonder})} & I wonder & who/if   &    & the & man & liked & ./him   & Wh   & No  & VP  & No  \\
Matrix Wh-Q                        &         & Who/""   & did& the & man & like  & ?/him   & Wh   & Yes & N/A & No  \\
Restr. Rel. Clause               & The boy & who/and  &    & the & man & liked & was/him & Wh   & No  & NP  & No  \\
Cleft                                     & It was  & the boy/clear & that & the & man & liked & ./the boy & Null & No  & VP  & Yes \\
Pseudo-Cleft                              &         & Who/That &    & the & man & liked & was/it & Wh   & No  & N/A & Yes \\
Topicalization                            & Actually, & the boy/"" &  & the & man & liked & ./the boy & Phrase & No & N/A & Yes \\
\midrule
Subject-Verb Agr.                    & The     & boy/boys & that & the & man & liked & is/are & \multicolumn{4}{c}{\textit{minimal non-filler-gap control}} \\
Trans/Intrans              &         & Once/Today & & some/that & man/boy & ran/liked & ./him & \multicolumn{4}{c}{\textit{lexically matched control}} \\
\bottomrule
\end{tabular}
}
\caption{\textbf{Left Block:} Exemplar minimal pairs for each construction’s single-clause, animate extraction variant (and controls). The filler/label combinations are used to evaluate whether the model is processing the construction correctly and whether our causal interventions are successful. \textsc{NC} (`no comparison') shows extra words required for the grammaticality of some constructions. As many don't require them, we do not train or test on them. For full sets of examples, including multi-clause and inanimate extraction variants, see \Cref{app:templates}.
\textbf{Right Block}: Parameters of linguistic variation for the same constructions. Our parameters (columns) are (1) nature of the filler (the class of the item which has `filled' the gap); (2) syntactic-head child inversion (whether the child of the constructions' syntactic-head has inverted to appear linearly before it); (3) syntactic category of the parent (the phrase under which the construction is embedded), and (4) the semantic/pragmatic nature of the construction (whether element fronting is done by syntactic necessity or for discourse purposes). Related regression results are in \Cref{tab:regression_results_exp2}.}
\label{tab:constructions_combined}
\end{table*}

\section{Filler--Gaps and Neural Models}

Consider the following sentence:

\begin{examples}
    \item $[$The bagel$]_t$, I liked \_\_$_t$.
    \label{ex:obj-extract3}
\end{examples}

\noindent The embedded clause, \textit{I liked}, seems incomplete, lacking an object. However, the sentence is grammatical, as the fronted entity \textit{the bagel} is understood to be the object of the anteceding clause.

Grammatical constructions of this nature are termed \textbf{filler--gaps}, due to
constituents appearing as `fillers' in non-canonical positions, colloquially being said to leave a `gap' at its canonical position. 
This grammatical family encompasses a wide range of common constructions including \textit{wh}-questions, relative clauses, clefts, and more. 

Filler--gap dependencies have long been a target of linguistic inquiry. They are believed to require sophisticated syntactic machinery, beyond simple surface statistics, since a word might appear quite linearly far from a word that it depends on for its meaning \citep{chomsky1957syntactic,ross1967constraints}.
They have been of interest in computational linguistics for the same reason: earlier models like n-gram models were fundamentally unable to handle structures over long distances.

Hence, filler--gaps have served as a common test-bed for LMs' grammatical capacities. \citet{wilcox-etal-2018-rnn} provided early positive evidence of RNNs' grammatical competence in English by comparing LMs' surprisals for gap and gapless continuations in the presence and absence of fillers. More recently \citet{ozaki2022well} and \citet{ wilcox2024using} have demonstrated LM sensitivities to linguistic constraints on these constructions. \citet{kobzeva-etal-2023-neural} found mixed results in Norwegian, a language known to have very different filler--gap structures and constraints than English. 

There has been further work to measure the generalization capacities of LMs across filler--gap constructions. \citet{lan2024large} test models' knowledge of parasitic gaps and across-the-board movement, finding that unless the training data is supplemented with adequate examples, LMs struggle to learn these constructions from small corpora. \citet{howitt-etal-2024-generalizations} build on the methodology of \citet{lan2024large}, training LSTMs on specific filler--gap constructions and evaluating LM performance on others, with results suggesting little generalization in LMs. \citet{prasad-etal-2019-using} and \citet{bhattacharya-van-schijndel-2020-filler} further use a methodology based on psycholinguistic priming to explore filler--gap generalization in LMs, with the former finding evidence suggesting that LMs hierarchically organize relative clauses in representation space, and the latter finding general representations for filler--gaps which are shared across various constructions. 

These previous works show LMs can learn to process filler--gap constructions, but show more mixed results as to whether this processing is shared across constructions. But most of this work has been behavioral, without exploring the model's underlying causal mechanisms. Our work fills this gap. We first uncover the causal mechanisms LMs learn to process various filler--gap dependencies, and then we measure to what extent these mechanisms generalize across different filler--gaps.

\section{Methods}

\subsection{Data}

\paragraph{Evaluated Constructions} We focus our investigation on seven filler--gap constructions: embedded wh-questions with a finite complementizer (denoted as the \textit{know}-class), embedded wh-questions with a non-finite complementizer (\textit{wonder}-class), matrix-level wh-questions, restrictive relative clauses, clefts, pseudo-clefts, and topicalization. For each construction, we design sentential templates in the style of \citet{arora-etal-2024-causalgym}, allowing us to sample a large number of minimal pairs differing in our targeted grammatical phenomenon.

We design four templates per construction, differing in the extracted object's animacy and by the number of clausal boundaries between the filler and the gap left by its extraction (one or two clauses). We manipulated animacy since changing animacy requires changing the key wh- element (``who'' vs. ``what''), but is not hypothesized to affect the sentence's structure. All our templates involve the extraction of a direct object from a verb phrase and all follow a general template, allowing cross-construction alignment by position. Our general template, as well as examples of animate extraction from a single-clause variant of each construction, can be found in Table \ref{tab:constructions_combined}. 

\paragraph{Controls} Our first control is the task of subject--verb number agreement (e.g., ``The boy is'', not ``The boy are''). 
This task was selected because, relative to our constructions of interest, there is a similar distance between the subject and the verb.
However, while subject--verb agreement can operate over long linear distances, it does not have the filler--gap property of our target constructions (as agreement is always between clausemate elements) and thus we hypothesize that it should \textit{not} rely on the same mechanism.

The second control is the task of predicting a continuation after transitive or intransitive verbs. This task controls for the predicted label,  ensuring that any generalization we find is meaningful, not merely due to heuristics related to the predicted labels. 
In order to maintain the distance between minimal contrast and prediction location, we have lexical items in faux-contrast at the \textsc{filler}, \textsc{article}, and \textsc{np} positions, such that there is no meaningful difference in the sampled items at those positions. 

\subsection{Distributed Alignment Search}\label{sec:DAS}

To localize internal mechanisms used by LMs to process our constructions of interest, we use Distributed Alignment Search (DAS; \citealt{wu2024interpretabilityscaleidentifyingcausal,geiger2024finding}). DAS is a supervised interpretability method that can be used to assess whether a given feature is encoded in a particular set of neural activations. We rely on the 1-dimensional variant of DAS used by \citet{arora-etal-2024-causalgym}. The core intervention performed is

\begin{equation*}
\setlength{\abovedisplayskip}{6pt} 
\setlength{\belowdisplayskip}{6pt}
\mathbf{b} + (\mathbf{s}\mathbf{a}^{\top} - \mathbf{b}\mathbf{a}^{\top})\mathbf{a}
\end{equation*}

\noindent where $\mathbf{b} \in \mathbb{R}^{n}$ is a representation formed by the model when it processes a base example (right sides in \Cref{fig:schematic}), and $\mathbf{s} \in \mathbb{R}^{n}$ is the corresponding representation formed when the model processes a source example (left sides in \Cref{fig:schematic}). In our experiments, $\mathbf{b}$ and $\mathbf{s}$ are always the outputs of a Transformer block. Intuitively, this intervention defines a direction in the rotated feature space defined by the learned vector $\mathbf{a} \in \mathbb{R}^{n}$. This is a soft intervention targeting only the learned feature and preserving orthogonal dimensions of $\mathbf{b}$. In DAS, all LM parameters are kept frozen, and $\mathbf{a}$ is learned via a standard cross-entropy loss trained on interventions of the sort depicted in \Cref{fig:schematic}. The goal of learning is to make the correct predictions under the intervention. For example, in the within-class intervention in \Cref{fig:schematic}, we seek to learn an intervention that predicts a gap site (signaled by a period) even though the inputs correspond to a non-filler--gap case. The extent to which we can learn such an intervention provides the basis for assessing the hypothesis that the filler--gap dependency itself can be localized to the intervention site.

We chose to use DAS for two main reasons. First, \citet{arora-etal-2024-causalgym} demonstrate that, in a comparison among several interpretability methods, DAS consistently performed the best in finding causally efficacious features in syntactic  tasks. Second, \citet{wu2024interpretabilityscaleidentifyingcausal} show that the feature-alignments learned by DAS are robust and generalize strongly.

\paragraph{Training} We train interventions at each position from the \textsc{filler} onwards, and across every layer of our given LM. We use the \texttt{pythia} series of models \citep{biderman2023pythiasuiteanalyzinglarge}, a series of open-source, open-data LMs. We run our experiments on the $1.4$, $2.8$ and $6.9$ billion parameter models. We find qualitatively similar results for all sizes, reporting those of the $1.4$b variant in the main text (results for $2.8$b and $6.9$b variants in \Cref{app:other-sizes}).

We train two distinct categories of interventions: (1) single-source interventions, where for each of the $n$ constructions, $c_{i < n}$, the training dataset for DAS contains sentences sampled from the templates of $c_i$, and (2) leave-one-out interventions, where for each of the $n$ constructions, $c_{i<n}$, the training dataset contains sentences sampled from the templates of $c_{j\not=i}$ -- that is, all constructions that are not `left-out'.\footnote{See \citet{rodriguez-etal-2025-characterizing} for a similar transfer approach to study semantic property inheritance in models.} In both settings, our training sets consist of 200 sentences sampled from the relevant constructions, before adding each sentence's minimal pair, resulting in perfectly balanced training sets of 400 sentences.

\paragraph{Evaluation}

For evaluation, we use the \textsc{\textbf{odds}} metric from \citet{arora-etal-2024-causalgym}. This metric measures how much more likely a counterfactual label (i.e., the mismatched word) is after performing an intervention, with higher \textsc{\textbf{odds}} denoting larger causal effect from the given intervention. 
Intuitively, it tells us: after intervention, how much more likely is the continuation expected based on the ``source sentence'' than the one naively expected based on the ``base sentence''. For each construction, we measure the average \textsc{\textbf{odds}} at each position-layer pair across 400 sentences, sampled so as to ensure no overlap with our training sets.

In cases of aggregation, we max-pool the average \textsc{\textbf{odds}} value across layers at each position (we refer to this metric as \textsc{\textbf{max odds}} hereafter). We also normalize the \textsc{\textbf{max odds}} by the corresponding average \textsc{\textbf{max odds}} for the items present in the training set. This normalization measures how much the mechanisms used by a given set of constructions generalize to an evaluated construction, relative to how much they generalize to those they were trained on. We aggregate across layers by max-pooling \textsc{\textbf{odds}} because our methodology aims to localize syntactic features in the model, with the maximum value representing the most causally efficacious localization of the given features.

\begin{figure*}
    \centering
    \includegraphics[width=1\linewidth]{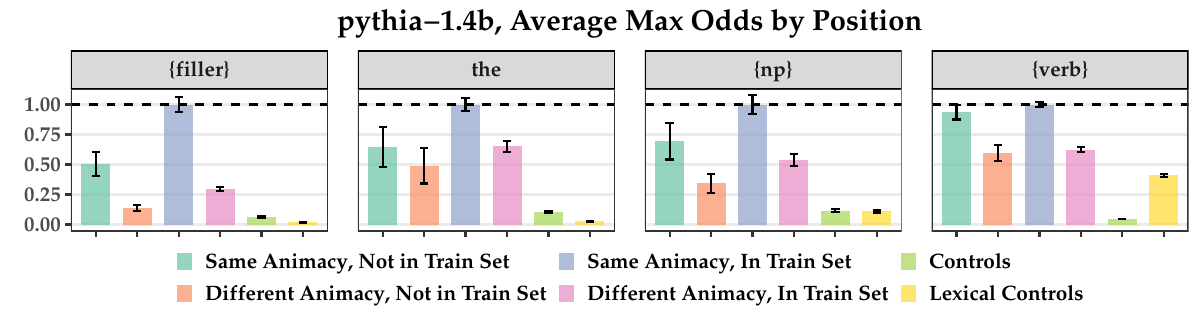}
    \caption{Average normalized \textsc{\textbf{max odds}} across positions, $\pm 1$ standard error. Corresponding multi-clause plots can be found in Appendix \ref{app:exp1-supp}. Note that normalization fixes the ``Same Animacy, In Train Set'' condition at 1.00.}
    \label{fig:loo-combined}
\end{figure*}

\begin{figure}
    \centering
    \includegraphics[width=.98\linewidth]{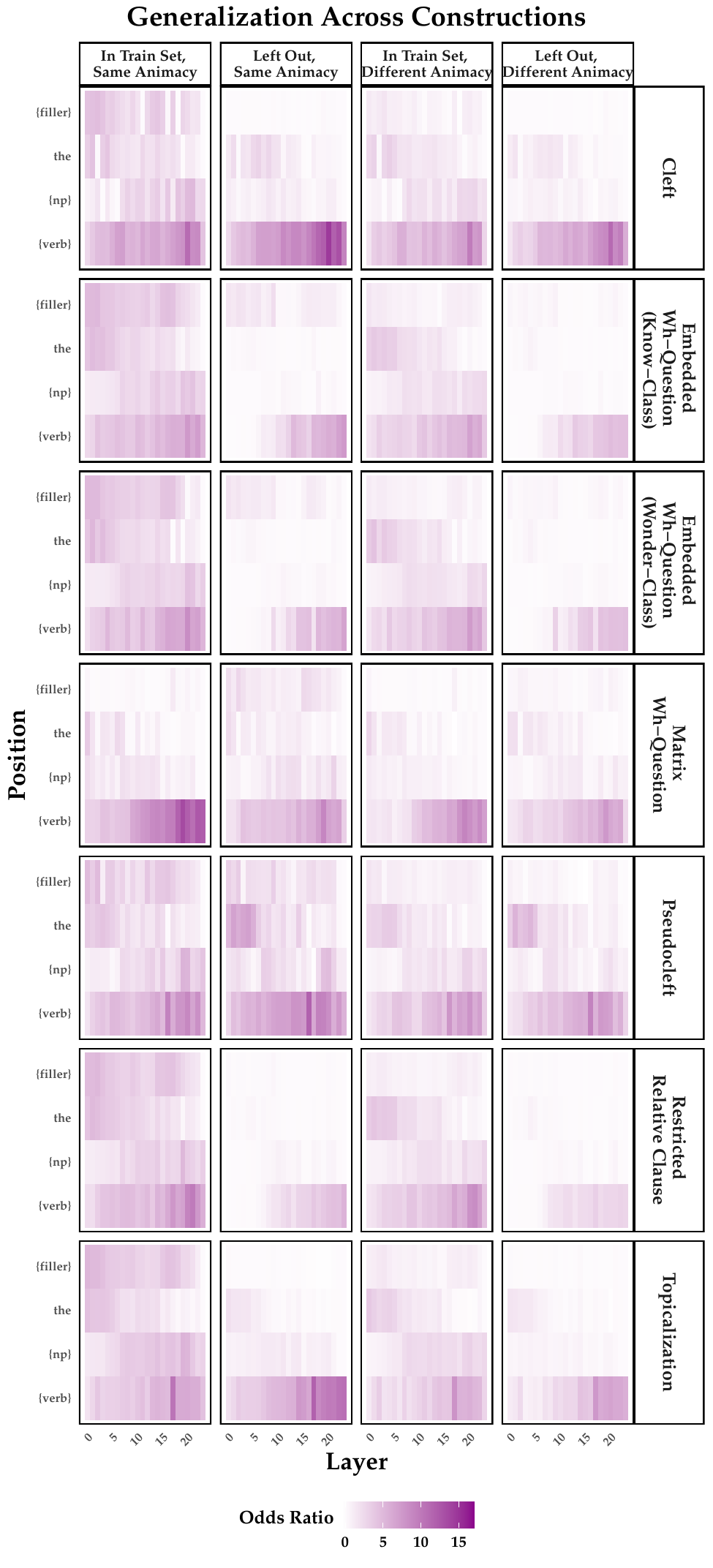}
    \caption{For each source construction, we measure the \textsc{\textbf{odds}} at each position--layer pair, aggregating the values by evaluation group. Corresponding plots with control values and multi-clause variants are in Appendix~\ref{app:exp1-supp}.}
    \label{fig:mech_loo}
\end{figure}

\section{Exp.~1: Do LMs Share Filler--Gap Mechanisms Across Constructions?}\label{sec:one}

Our first experiment investigates the extent to which language models employ common mechanisms for processing different filler--gaps. 

\paragraph{Setup} We measure the \textsc{\textbf{max odds}} for all trained interventions evaluated on every construction of the same clausal category (for a discussion on cross-clause generalization see \Cref{sec:cross-clause}). We then group these values into six categories, based on the relation between the set of constructions the interventions were trained on and those used to generate the evaluation set. These groups comprise (1) the same set of constructions in the training set and the evaluation set, with the same animacy -- this is our reference group as training and evaluation sentences are drawn from the same distribution; (2) the same set of constructions in training set and evaluation set, with different animacy; (3) evaluation on the held-out constructions, but with the same animacy as the training set; (4) evaluation on the held-out constructions, and differing animacy from the training set; and (5--6) the two controls. Conditions (1) and (2) are examples of the `single source interventions' as described in \Cref{sec:DAS}, with the single construction present in the training set also being present in the evaluation set, with (3) and (4) examples of `leave-one-out interventions', with training sets including all constructions except the one present in the evaluation set. The sole difference between the items within these pairs is whether the animacy conditions match in the training and evaluation sets.

\paragraph{Hypothesis} We hypothesize that the \textsc{\textbf{max odds}} for all our targeted evaluation groups will be greater than that of the controls. 
We further expect  \textsc{\textbf{max odds}} to be higher when the evaluated constructions are in the training set or match in animacy.

\paragraph{Results} 
\Cref{fig:loo-combined} shows the average \textsc{\textbf{max odds}} of the aforementioned groups at each position in our single-clause templates. In both these single-clause variants and the multi-clause variants of our constructions (corresponding figure in \Cref{app:exp1-supp}), we find consistently high \textsc{\textbf{max odds}} values for each of the non-control groups. The controls show significantly less transfer.
We run pairwise t-tests with a Holm-Bonferroni correction, finding the \textsc{\textbf{max odds}} of each of our test groups is significantly higher than both controls at every position in the single-clause templates and nearly every position in the multi-clause ones. These results strongly suggest \textbf{shared internal representations across filler--gap constructions in the evaluated models.}

To test our hypotheses regarding the effect of training and evaluation set overlap and matching animacy, we fit a linear mixed-effects regression model to our \textsc{\textbf{max odds}} data at each position. Our random effects are intervention training set and evaluation construction, and our fixed effects take the form of binary indicator variables for (1) whether the evaluated construction was in the training set and (2) whether the animacy condition of the evaluated construction matches that of the training set. We find significant, positive, effects for overlap, matching animacy, and their interaction at the \textsc{filler}, \textsc{the}, and \textsc{np} positions, and for matching animacy at the \textsc{verb} position. See \Cref{app:exp1reg} for regression details. 
Thus, across positions, \textbf{LM internal processing is sensitive to linguistically meaningful features, such as animacy of the extracted item} (possible evidence of ``lexical boost'').

While we broadly see generalization as fitting into held-out constructions (\Cref{fig:mech_loo}), embedded wh- questions and restrictive relative clauses show noticeably less generalization than other constructions. We briefly offer up two accounts for these peculiarities: (1) there is asymmetry in LM generalization between different filler--gap dependencies or (2) these constructions are processed by largely different mechanisms than the other constructions. Clarifying which of these applies to each construction helps motivate our next experiment.

\section{Exp.~2: What Factors Drive Filler--Gap Generalization in LMs?}\label{sec:factors}

\noindent 
Our previous experiment demonstrated significant overlap between the LM's abstract representations of various filler--gap constructions. However, we also observed notable variation in the strength of this generalization across positions and constructions. Here, we aim to characterize the nature of this cross-construction generalization. In particular, we aim to identify whether there exist constructions which serve as \textit{sources} (their filler--gap properties transfer well to other constructions) or \textit{sinks} (filler--gap properties from other constructions transfer well to them). We further investigate which features of natural language (e.g.\ distributional properties like frequency, or linguistic properties like the nature of the filler item) drive this generalization.

\begin{figure}[t]
    \centering
    
    \begin{subfigure}[t]{1\linewidth}
        \centering
        \includegraphics[width=\linewidth]{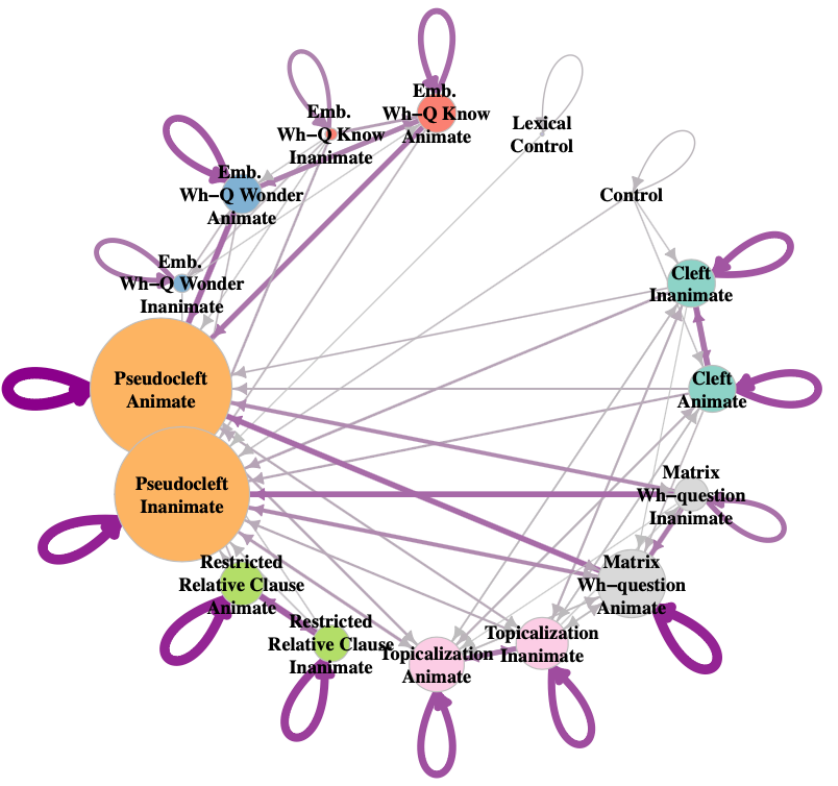}
    \end{subfigure}%
    \hfill
    \begin{subfigure}[t]{0.9\linewidth}
        \centering
        \includegraphics[width=\linewidth]{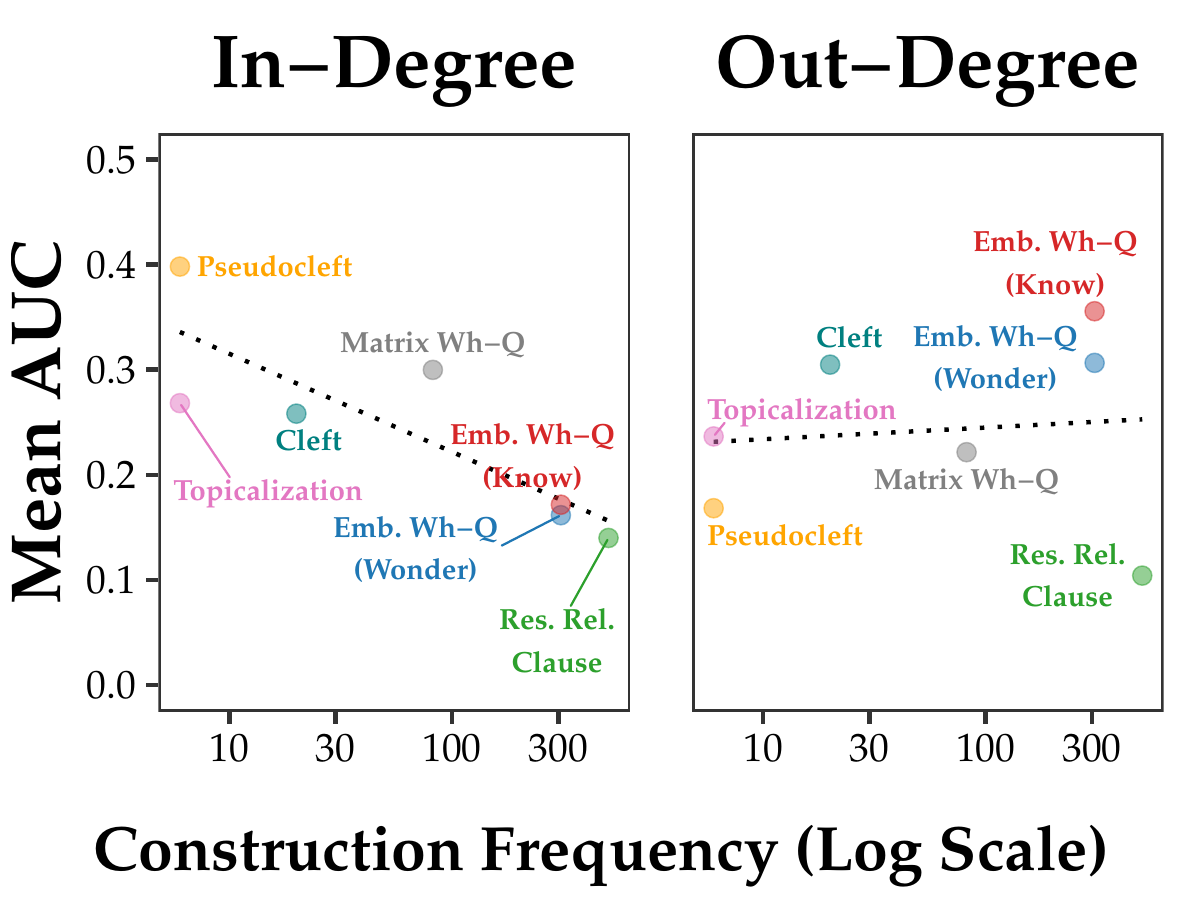}
    \end{subfigure}%
    \caption{\textbf{Top}: Generalization network at single-clause \textsc{the} position with edge-threshold of 1. Node size proportional to in-degree; edge size and color proportional to \textsc{\textbf{odds}} of the source construction's interventions measured on the target construction. \textbf{Bottom:} In- and out-degree centrality AUCs against construction frequency. }
    \label{fig:source-sink-scatter}
\end{figure}

\paragraph{Setup} To characterize the degree to which a given construction is a source or sink, we perform the following procedure. First, we evaluate all single-source interventions on all constructions of the same clausal length, averaging the normalized \textbf{\textsc{max odds}} across the animacy-conditions at each position, training construction, and evaluation construction triple. At each position, we take the resulting $n \times n$ matrix to be an adjacency matrix for a weighted, directed graph $G = (V,E)$ in which vertex $V$ is a construction and each directed edge $E_{i,j}$ is the transfer from construction $i$ to construction $j$. We then calculate \textit{out-degree centrality} -- the fraction of a graph's total nodes that a given node's outgoing edges are connected to -- and \textit{in-degree centrality} -- the fraction of nodes its incoming edges come from.
We do this across a range of edge thresholds -- i.e., the minimum edge weight retained in the graph.
We measure each construction's area under the threshold-centrality curves (AUC). 
The resulting out- and in-degree AUCs serve as proxies for the degree to which a given construction is a source or sink respectively. We provide an exemplar generalization network (for the \textsc{the} position) in  \Cref{fig:source-sink-scatter}.
That figure shows particularly strong transfer into pseudo-clefts, very little transfer into either control, strong within-construction transfer (dark recurrent arrows), and some non-random structure of transfer across constructions.

We also analyzed the effect of construction frequency on generalization capacity. We extracted
estimates of each construction's prevalence in the English-EWT Universal Dependencies dataset \citep{de2021universal, nivre-etal-2020-universal, silveira-etal-2014-gold}. See \Cref{sec:appendix-UD} for details.%

We further investigate the effects of four parameters of linguistic variation across filler--gap constructions: the nature of the filler, whether the head child is inverted, the syntactic category of the parent (the word under which a construction is embedded), and the semantic/pragmatic nature of the construction (whether the fronted element is fronted by necessity or for discourse reasons), with \Cref{tab:constructions_combined} presenting the associated parameter value for each construction. At each position, we fit a linear mixed-effects model predicting the \textbf{\textsc{max odds}}, with binary indicator variables denoting whether the source and evaluated construction match for each of the above posited parameters of variation as fixed effects, with random effects for training-source construction, and evaluated construction. For full regression details, see  \Cref{app:exp2reg}.

Finally, we perform Principal Component Analysis (PCA) at each position, reducing the dimensionality of our generalization matrix to the two principal components, allowing visualization of construction similarity in this space.

\paragraph{Hypothesis} We expect some constructions to serve as strong sources and others as strong sinks in the generalization network. We further expect a positive relationship between a construction's frequency and the degree to which it is a source, and conversely, a negative relationship between its frequency and its sink-ness.
Finally, we anticipate stronger generalization between linguistically similar constructions than dissimilar ones (as operationalized by the parameters of linguistic variation in \Cref{tab:constructions_combined}).

\paragraph{Results} 

\begin{figure}
    \centering
    \includegraphics[width=1\linewidth]{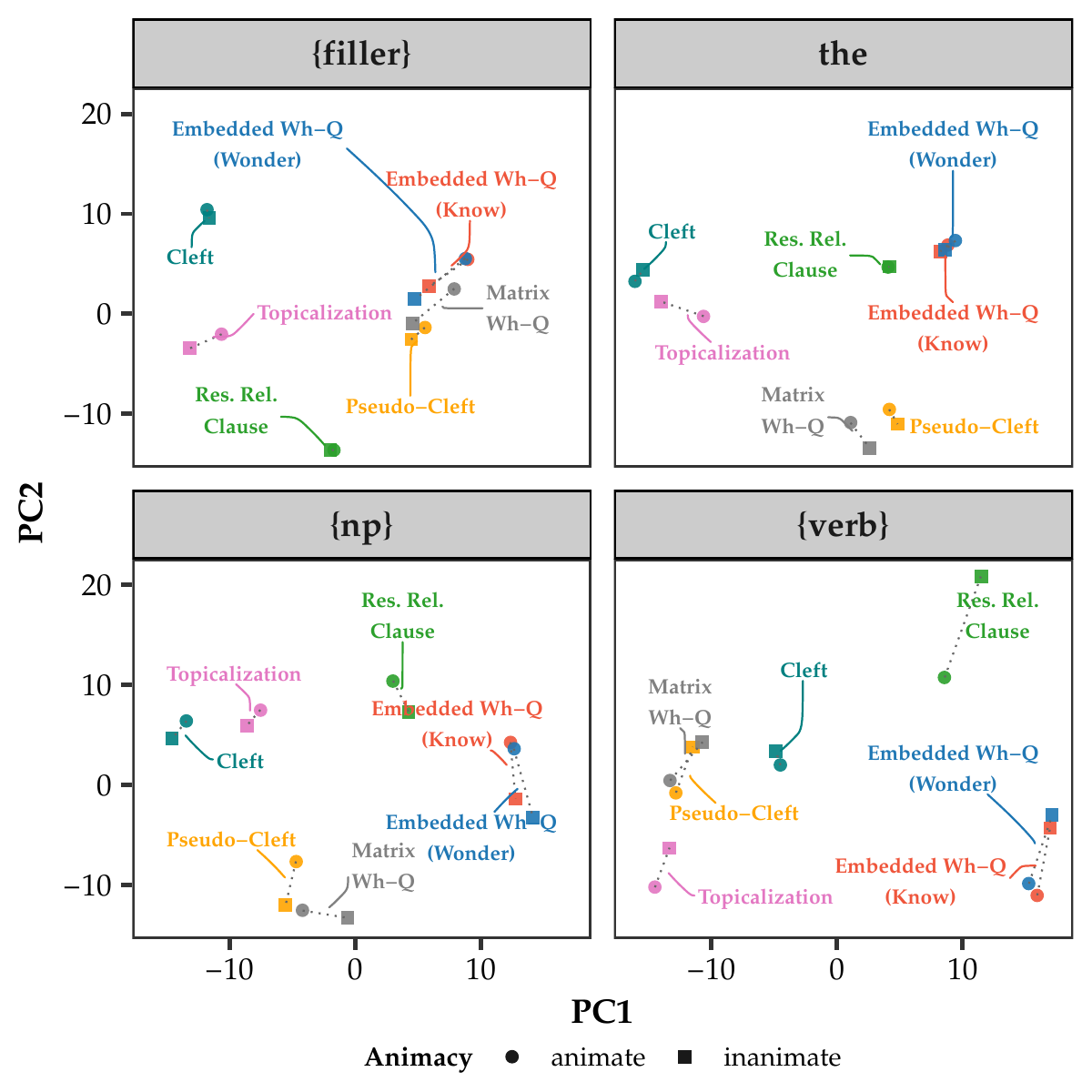}
    \caption{Constructions plotted along the top two principal components at each position in our single-clause variants. Generally, constructions cluster in linguistically intuitive ways -- e.g. animate/inanimate pairs generally cluster, constructions with wh-fillers cluster at the \textsc{filler} position, and restrictive relative clauses typically lie away from the other analyzed constructions.}
    \label{fig:pca}
\end{figure}

\Cref{fig:source-sink-scatter} shows construction frequency against in-degree and out-degree AUCs, mean-pooled across sentence positions. Constructions are spread across the AUC axis, suggesting varying levels of generalization. These AUCs are consistent across both sentence position and clausal variant (single and multi-clause AUCs, faceted by position, are available in \Cref{app:exp2-supp}). 

Figure \ref{fig:source-sink-scatter} also shows a negative relationship between construction frequency and in-degree AUC and a (weak) positive relationship between construction frequency and out-degree AUC. There are some notable exceptions to these trends, such as the low-frequency topicalization construction having a surprisingly low in-degree AUC and the most frequent construction, restrictive relative clauses, having a low out-degree AUC. 
Below, we argue that these anomalies are linguistically explainable.

\begin{table}[t]
\centering
\footnotesize
\begin{tabular}{lllll}
\toprule
\textbf{Term} & \textbf{$\beta$}$_{\textsc{filler}}$ &  \textbf{$\beta$}$_{\textsc{the}}$ &  \textbf{$\beta$}$_{\textsc{np}}$ &  \textbf{$\beta$}$_{\textsc{verb}}$ \\
\midrule
(Intercept)                  & \textbf{1.15***} & \textbf{1.96***} & \textbf{1.32***} & \textbf{7.12***}  \\
\texttt{filler}      & \textbf{0.75***} & \textbf{1.06**} & 0.28     & 0.53  \\
\texttt{inverted}       &   \textbf{0.38**} & \textbf{0.51**} & \textbf{0.40**}     & 0.06 \\
\texttt{embed} &   \textbf{0.85***} & \textbf{1.05**} & \textbf{0.54**} & \textbf{2.06**}  \\
\texttt{front} & \textbf{0.30**} & \textbf{0.36*} & \textbf{0.34***} & 0.32  \\
\bottomrule
\end{tabular}
\caption{\textbf{Experiment 2 Regression Results.} * denotes $p<.05$, ** denotes $p<.01$, and *** denotes $p<.001$. The dependent variable is the \textbf{\textsc{max odds}} 
The coefficients correspond to the linguistic variables of interest shown in Table \ref{tab:constructions_combined}. Generalization tends to be significantly greater when linguistic properties are shared.
}
\label{tab:regression_results_exp2}
\end{table}

We further find evidence supporting our hypothesis that linguistic similarity aids generalization between constructions. Our regression (Table \ref{tab:regression_results_exp2}) reveals significant, positive effects for filler type at the \textsc{filler} and \textsc{the} positions, inversion of the head child and nature of the fronted element at the \textsc{filler}, \textsc{the}, and \textsc{np} positions, and syntactic category of the parent at all positions. Furthermore, Figure \ref{fig:pca} demonstrates convergent results from PCA, with linguistically related constructions generally clustering along the principal components. For instance, animate and inanimate forms of the same construction tend to cluster together, and cleft and topicalization constructions tend to cluster together.

\paragraph{Discussion}

These results paint a clear picture of filler--gap generalization in LMs. Frequent constructions are encountered at a high-enough rate during training to drive the development of robust mechanisms to process them. Less frequent constructions are not encountered enough for stand-alone, robust processing mechanisms to form. Instead, their processing relies on the mechanisms of more frequent, linguistically similar constructions. 

Further linguistic analyses reveal effects beyond frequency.
For instance, we observed a low in-degree AUC for the low-frequency construction topicalization. Topicalization is linguistically dissimilar to higher-frequency constructions, being the only construction with a phrasal element at its filler site, and it generally shares very few linguistic features with more frequent constructions. In this light, its low in-degree AUC is not surprising, especially when compared to pseudo-clefts, which much more closely resemble higher-frequency constructions (especially wh-questions).

Similarly, restrictive relative clauses are the only constructions which are embedded under a noun phrase, possess a wh-item at the filler position, and have their filler item fronted out of syntactic necessity, not for discourse purposes. This makes them linguistically dissimilar to many of the lower frequency constructions along the features found important by our mixed-effects model. As such, despite their high frequency, their mechanisms do not transfer broadly to these constructions, leading to a relatively low out-degree. 

These results also answer the questions posed at the end of Experiment 1. Namely, embedded wh-questions and restrictive relative clauses show little generalization in the leave-one-out setting, as they are frequent enough to largely not rely on the processing mechanisms of other constructions. However, embedded wh-questions possess enough linguistic overlap with less frequent constructions to aid in their processing, whereas restrictive relative clauses are more isolated in the generalization network due to their linguistic dissimilarities.

\begin{figure*}
    \centering
    \includegraphics[width=1\linewidth]{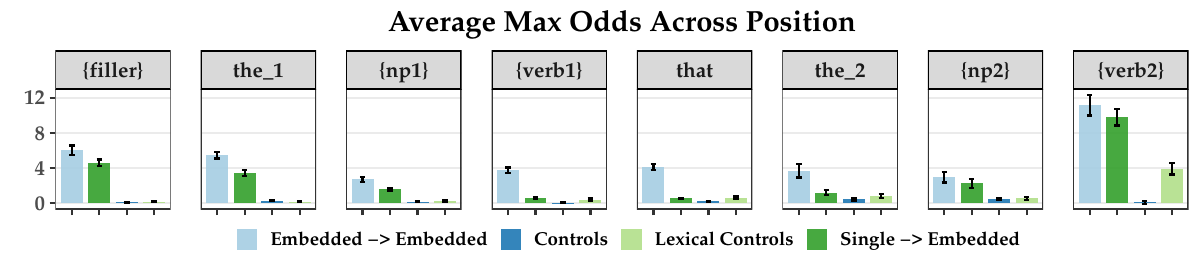}
    \caption{\textsc{\textbf{max odds}} $\pm 1$ standard error, by position for interventions (1) trained and evaluated on multi-clause variants, (2) trained on single-clause variants and evaluated on multi-clause variants, and (3--4) controls. 
    }
    \label{fig:sd}
\end{figure*}

\section{Exp.~3: Do Language Models Generalize Across Clausal Boundaries?}\label{sec:cross-clause} 

\noindent Our first two experiments demonstrate that LMs share processing mechanisms across various filler--gap constructions of the same clausal length. In this section, we analyze whether our constructions' single-clause processing mechanisms are used to process both clauses in the multi-clause variant.

\paragraph{Setup} We evaluate the interventions trained at each position of the single-clause variants on the corresponding positions in the matrix and embedded clause of the same construction's multi-clause template. We compare the by-position \textsc{\textbf{max odds}} values to the corresponding values of interventions trained and evaluated on the multi-clause variants.

\paragraph{Hypothesis} Under a purely modular account of syntactic structure, we expect to see generalization across clausal boundaries. That is, we expect the single-clause interventions to show above-chance \textsc{\textbf{max odds}} when evaluated on both the matrix and embedded clause of our multi-clause variants.

\paragraph{Results} Our results are displayed in \Cref{fig:sd}. Overall, while we see robust transfer between embedded clauses, we see little meaningful transfer from matrix to embedded clauses.

Our single-clause mechanisms show above-chance \textsc{\textbf{max odds}} at the \textsc{filler} through \textsc{np}$_1$ positions of the matrix clause, before dropping off at the \textsc{verb}$_1$ through \textsc{the}$_2$ positions, and then slowly rebounding as we move towards the final \textsc{verb}$_2$. 

These results make sense when we consider the relative sentential structures of single-clause and multi-clause sentences, and the auto-regressive nature of the LMs we study. The first three positions of a multi-clause sentence -- that is, \textsc{filler}, \textsc{the}$_1$, and \textsc{np}$_1$ -- are indistinguishable from the first three positions of a singular-clause sentence. As such, we would expect an auto-regressive LM, processing from left-to-right, to not be aware that it is processing an embedded clause until it reached the \textsc{verb}$_1$ position. Until then, it will use the same mechanisms it would to process a sentence with a single clause. This is reflected in the strong generalization through these first three positions.

In the \textsc{verb}$_1$ position, however, single-clause and multi-clause sentences have verbs that sharply diverge in their semantic character and syntactic properties. Specifically, the verbs at this position in a multi-clause sentence must be ones which can embed a clause (e.g.\ \textit{say}, \textit{know}, and \textit{wonder}, among others), whereas in a single-clause sentence this is not necessary. As such, upon encountering this position, the LM encounters a different set of verbs than it was trained on, leading to a drop in the single-clause intervention's \textsc{\textbf{max odds}}. 

As the LM processes the next couple of positions (\textsc{that}, \textsc{the}$_2$, and \textsc{np}$_2$), we see the single-clause intervention's \textsc{\textbf{max odds}} steadily increasing, as the LM gets closer to a position where it can potentially discharge its filler. This process culminates at the \textsc{verb}$_2$ where we see clear, above-chance, generalization from the single-clause mechanisms to the embedded-clause.

\paragraph{Discussion} 
While many syntactic theories posit that filler-gap structures are processed uniformly across contexts, our findings suggest that, in LMs, filler-gap constructions are handled by different mechanisms in matrix and embedded clauses.

\section{Conclusion}

Long-held views in linguistics suggest that there should be common processing characteristics across diverse English filler--gap constructions. 
We found this largely to be the case for LMs: we were able to transfer the filler--gap property across neural representations of different filler--gap constructions, suggesting that neural models rely on similar representations across distinct constructions.

Moreover, our analyses suggest that the strength of the transfer is mediated by linguistically interesting properties. We see a significant, positive boost in generalization for constructions matching in filler type, for constructions with similar verb inversion patterns, for constructions that match in whether they involve information-structural fronting, and for constructions in which the relevant syntactic parents share a syntactic category.

The transfer effects are not uncomplicated, though. The observed structural effects are accompanied by frequency and animacy effects. The processing mechanisms of more frequent constructions support the processing of less frequent constructions.
And, even across constructions which are syntactically \textit{identical} but differ in animacy, transfer is weaker than when animacy matches.
This was true even though animacy and frequency are not a key part of the usual syntactic account of filler--gap constructions.
We also found transfer between embedded and matrix clauses to be weak.

These findings point to linguistically interesting hypotheses about the factors governing constructional similarity -- hypotheses that could directly inform future linguistic research by, for instance, exploring whether humans show greater priming effects across filler--gap constructions that share these relevant properties. We argue that mechanistic analysis of LMs can provide novel insights into the nature of syntactic structures.

\section{Limitations}

Our work is primarily an attempt to show that LMs can be useful tools for pushing linguistic theory forward. This brings with it specific theoretical presuppositions that are worth articulating to avoid a suggestion that there is scientific consensus where there is not.

Our investigation is oriented toward finding evidence of modular structure in LMs. However, it is not a settled question what constitutes rule-like or systematic linguistic behavior in neural systems \citep{nefdt2023language,geiger2024finding,buckner2024deep,futrell2025linguistics}. How causally systematic should a syntactic behavior be for it to be rule-like? One reading of our results would be that our causal interventions capture human filler--gap behavior but noisily (e.g., imperfect transfer across constructions, less transfer when animacy differs).

This is possible, but another reasonable interpretation is that the relevant constructs are also fuzzy in humans.
Despite a historical proclivity for rules, nearly all syntactic theories allow for numerous exceptions, and human behavior itself is variable and subject to errors. As such, the questions we ask regarding the rule-like nature of LMs extend beyond such models, becoming broader questions about human processing and behavior. Our findings alone cannot adjudicate these questions, though.

Further, while we provide evidence in \Cref{sec:factors} that the frequency of a given construction plays a large role in its strength as a source, we do not preclude that this is the only factor driving source strength. Specifically, we do not rule out that the inductive biases present in Transformer-based LMs may inherently process certain constructions better than others. However, our study is not designed in a manner such that we can address this, and we leave this as a direction for future work.

We also note that our results are only in English. It would be valuable to extend them to other languages, particularly those with typologically different filler--gap patterns.

We relied here on templatically generated sentences, which are known to differ in systematic ways from naturally occurring sentences. We would like to extend this work to naturalistic sentences, but doing so is challenging because of the strong constraint that we have matched pairs.

\section*{Acknowledgments} 

We would like to thank Qing Yao and, more broadly, the whole computational linguistics research group at UT Austin for their helpful conversations regarding this project. We thank audiences at Saarland University, Edinburgh University, and Dagstuhl Seminar 25301 for helpful comments. We acknowledge funding from NSF CAREER grant 2339729 to Kyle Mahowald and from Google and Open Philanthropy to Christopher Potts.

\bibliography{anthology1,anthology2,anthology3,custom,everything}

\appendix

\section{Construction Templates}\label{app:templates}

We provide templates and examples for our single-clause inanimate extraction (\Cref{tab:single-inanimate}), multi-clause animate extraction (\Cref{tab:multi-animate}), and multi-clause inanimate extractions (\Cref{tab:multi-inanimate}). In these tables, we use the shorthand demonstrated in \Cref{tab:abbreviations} to refer to our constructions.

\begin{table}[htbp]
\centering
\resizebox{\columnwidth}{!}{%
\begin{tabular}{@{}lll@{}}
    \toprule
    \textbf{Full Construction} & & \textbf{Shorthand} \\
    \midrule
    Emb. Wh-Question (\textit{Know}-Class) & & Emb. Wh-Q (\textit{K}) \\
    Emb. Wh-Question (\textit{Wonder}-Class) & & Emb. Wh-Q (\textit{W}) \\
    Matrix Wh-Question & & Matrix Wh-Q \\
    Restrictive Relative Clause & & RRC \\
    Pseudo-Cleft & & PC \\
    Topicalization & & Topic \\
    \midrule
    Subject-Verb Agreement & & SVA \\
    Transitive/Intransitive Verbs & & T/I Verbs \\
    \bottomrule
\end{tabular}%
}
\caption{Abbreviations for syntactic constructions in \cref{tab:single-inanimate,tab:multi-animate,tab:multi-inanimate}.}
\label{tab:abbreviations}
\end{table}

\begin{table*}[htbp]
    \centering
    \resizebox{\textwidth}{!}{%
        \small
        \begin{tabular}{llllllll}
            \toprule
            \textbf{Construction} & \textbf{Prefix} & \textbf{Filler} & \textbf{NC} & \textbf{Article} & \textbf{NP} & \textbf{Verb} & \textbf{Label} \\
            \midrule
            Emb.~Wh-Q (\textit{K}) & I know & what/that & & the & man & built  & ./it  \\
            Emb.~Wh-Q (\textit{W}) & I wonder & what/if & & the & man & built & ./it \\
            Matrix Wh-Q & & What/"" & did & the & man & build & ?/it \\
            RRC & The chair & which/and & & the & man & built & was/it\\
            Cleft & It was & the chair/clear & that & the & man & built & ./the chair\\
            PC & & What/That & & the & man & built & was/it\\
            Topic. & Actually, & the chair/"" & & the & man & built & ./the chair\\
            \midrule
            SVA & The & boy/boys & that & the & man & liked & is/are \\
            T/I Verbs & & Last night/Yesterday & & some/that & man/boy & ran/built & ./it \\
            \bottomrule
        \end{tabular}
    }
    \caption{Template and exemplar sentences for inanimate extraction from our single-clause construction variants.}
    \label{tab:single-inanimate}
\end{table*}

\begin{table*}[htbp]
    \resizebox{\textwidth}{!}{%
        \begin{tabular}{llllllllllll}
            \toprule
            \textbf{Construction} & \textbf{Prefix} & \textbf{Filler} & \textbf{NC} & \textbf{Article$_1$} & \textbf{NP$_1$} & \textbf{Verb$_1$} & \textbf{that} & \textbf{Article$_2$} & \textbf{NP$_2$} & \textbf{Verb$_2$} & \textbf{Label} \\
            \midrule
            Emb.~Wh-Q (\textit{K}) & I know & who/that & & the & nurse & said & that & the & man & liked & ./it  \\
            Emb.~Wh-Q (\textit{W}) & I wonder & who/if & & the & nurse & said & that & the & man & liked & ./it \\
            Matrix Wh-Q & & Who/"" & did & the & nurse & say & that & the & man & liked &  ?/it \\
            RRC & The boy & who/and & & the & nurse & said & that & the & man & liked &  was/it\\
            Cleft & It was & the boy/clear & that & the & nurse & said & that & the & man & liked &  ./the chair\\
            PC & & Who/That & & the & nurse & said & that & the & man & liked &  was/it\\
            Topic. & Actually, & the boy/"" & & the & nurse & said & that & the & man & liked & ./the chair\\
            \midrule
            SVA & The & boy/boys & that & the & nurse & said & that & the & man & liked &  is/are \\
            T/I Verbs & & Last night/Yesterday & & the & nurse & said & that & some/that & man/boy & ran/liked &  ./it \\
            \bottomrule
        \end{tabular}
    }
    \caption{Template and exemplar sentences for animate extraction from our multi-clause construction variants.}
    \label{tab:multi-animate}
\end{table*}

\begin{table*}[htbp]
    \resizebox{\textwidth}{!}{%
        \begin{tabular}{llllllllllll}
            \toprule
            \textbf{Construction} & \textbf{Prefix} & \textbf{Filler} & \textbf{NC} & \textbf{Article$_1$} & \textbf{NP$_1$} & \textbf{Verb$_1$} & \textbf{that} & \textbf{Article$_2$} & \textbf{NP$_2$} & \textbf{Verb$_2$} & \textbf{Label} \\
            \midrule
            Emb.~Wh-Q (\textit{K}) & I know & what/that & & the & nurse & said & that & the & man & built & ./it  \\
            Emb.~Wh-Q (\textit{W}) & I wonder & what/if & & the & nurse & said & that & the & man & built & ./it \\
            Matrix Wh-Q & & What/"" & did & the & nurse & say & that & the & man & built &  ?/it \\
            RRC & The chair & which/and & & the & nurse & said & that & the & man & built &  was/it\\
            Cleft & It was & the chair/clear & that & the & nurse & said & that & the & man & built &  ./the chair\\
            PC & & What/That & & the & nurse & said & that & the & man & built &  was/it\\
            Topic. & Actually, & the chair/"" & & the & nurse & said & that & the & man & built & ./the chair\\
            \midrule
            SVA & The & boy/boys & that & the & nurse & said & that & the & man & liked & is/are \\
            T/I Verbs & & Last night/Yesterday & & the & nurse & said & that & some/that & man/boy & ran/built & ./it \\
            \bottomrule
        \end{tabular}
    }  
    \caption{Template and exemplar sentences for inanimate extraction from our multi-clause construction variants.}
    \label{tab:multi-inanimate}
\end{table*}

\section{Training and Evaluation Details}\label{app:train-eval-detail}

We access the \texttt{pythia} models used in this study through the \texttt{transformers} python package \citep{wolf2020huggingfacestransformersstateoftheartnatural}. To train DAS, we use the \texttt{pyvene} library \citep{wu-etal-2024-pyvene} and follow the hyperparameters used by \citet{arora-etal-2024-causalgym}.

Our evaluation sets for the \texttt{pythia-1.4b} models consist of 400 sentences, with \textsc{\textbf{odds}} at each position-layer pair averaged across all evaluation sentences. For the other model variants evaluated (\texttt{pythia-2.8b} and \texttt{pythia-6.9b}) we use evaluation sets of 96 sentences due to computational constraints, noting that this is still larger than the prescribed evaluation size of 50 sentences from \citet{arora-etal-2024-causalgym}. We ensure that the intersect of train sets and evaluation sets is empty, so as to not bias our evaluations. Our training and evaluation ran on 2 NVIDIA A40 GPUs. For one model size, training totaled $\approx$12 hours, and evaluation $\approx$250 hours.

\begin{figure*}[!bp]
    \centering
    \begin{align*}
        \texttt{model <- lmer(max$\_$odds)} \sim  &\texttt{(1 + \texttt{in$\_$train$\_$set * same$\_$animacy} | from) + } \\
        &\texttt{(1 + \texttt{in$\_$train$\_$set * same$\_$animacy} | to) + } \\
        & \texttt{in$\_$train$\_$set * same$\_$animacy}
    \end{align*}
    \caption{Model formula used at each position for the linear mixed-effects regressions in Experiment 1.}
    \label{fig:reg1}
\end{figure*}

\begin{figure*}[!bp]
    \centering
    \begin{align*}
        \texttt{model <- lmer(max$\_$odds)} \sim &\texttt{(1 + match$\_$filler$\_$class + match$\_$inversion + }\\ &\texttt{match$\_$embedded$\_$under + match$\_$discourse$\_$fronted || from ) + }\\
        &\texttt{(1 + match$\_$filler$\_$class + match$\_$inversion + }\\ &\texttt{match$\_$embedded$\_$under + match$\_$discourse$\_$fronted || to ) + } \\
        & \texttt{match$\_$filler$\_$class + match$\_$inversion + }\\
        & \texttt{match$\_$embedded$\_$under + match$\_$discourse$\_$fronted}
    \end{align*}
    \caption{Model formula used at each position for the linear mixed-effects regressions in Experiment 2.}
    \label{fig:reg2}
\end{figure*}

\section{Regression Details}\label{app:regression}

We perform all regressions with the \texttt{lmerTest} package in R \citep{kuznetsova2017lmertest}. 

\subsection{Experiment 1 Regression}\label{app:exp1reg}

In the leave-one-out setting, we fit a linear mixed-effects model at each position with our dependent variable as the \textsc{\textbf{max odds}} at each training -- evaluation-set pair. We treat the training-set and evaluation-set as random effects, with fixed effects comprising indicator variables for whether the constructions in the training-set and evaluation-set match and whether animacy of the training-set and evaluation-set match. We also include a term for their interaction. As per \citet{barr2013random}, we include maximal random effect slope structures. Our full regression model is as in \Cref{fig:reg1}, which we fit to obtain the reported $\beta$ coefficients, and corresponding p-values.

Indicator variables are codified such that if the evaluated construction is in the training-set, $\texttt{in$\_$train$\_$set = 1}$ with $\texttt{in$\_$train$\_$set = -1}$ otherwise. Similarly, if the evaluated construction's animacy matches that of the training conditions, $\texttt{same$\_$animacy = 1}$ with $\texttt{same$\_$animacy = -1}$ otherwise. Table \ref{tab:regression_results_exp1} shows full regression results. \textit{Note: In this setting, the \texttt{construction$\_$from} variable denotes the held-out construction.} 

\begin{table*}[ht]
\centering
\begin{tabular}{lrrrr}
\toprule
\textbf{Term} & \textbf{$\beta$}$_{\textsc{filler}}$ &  \textbf{$\beta$}$_{\textsc{the}}$ &  \textbf{$\beta$}$_{\textsc{np}}$ &  \textbf{$\beta$}$_{\textsc{verb}}$ \\
\midrule
(Intercept)                  & \textbf{1.93***} & \textbf{2.70***} & \textbf{1.87***} & \textbf{9.06***}  \\
\texttt{in\_train\_set}      & \textbf{0.67***} & \textbf{0.56***} & \textbf{0.42**} & 0.26  \\
\texttt{same\_animacy}       & \textbf{1.08***} & \textbf{0.51***} & \textbf{0.60***} & \textbf{2.13***}  \\
\texttt{in\_train\_set:same\_animacy} & \textbf{0.36**} & \textbf{0.20**} & \textbf{0.10*} & 0.10  \\
\bottomrule
\end{tabular}
\caption{Experiment 1 Regression Results. * denotes $p<.05$, ** denotes $p<.01$, and *** denotes $p<.001$. The dependent variable is the \textbf{\textsc{Max Odds}}. Coefficients correspond to relationship between the training set and evaluation set for a given intervention. Generalization tends to be stronger when the evaluated constructions are in the training set and match in animacy.}
\label{tab:regression_results_exp1}
\end{table*}

\subsection{Experiment 2 Regression}\label{app:exp2reg}

In the single-construction setting, we fit a linear mixed-effects model at each position with our dependent variable as the \textsc{\textbf{max odds}} at each training-set and evaluation-set pair. We treat the training-set and evaluation-set as random effects. Our mixed-effects comprise indicator variables denoting whether the training construction and the evaluation construction match in our proposed filler--gap parameters of variation. A full breakdown of these parameters of variation and how they apply to our constructions of interest can be seen in Table \ref{tab:constructions_combined}. The resulting indicator variables take a value of 1 if the construction in the trainset and the construction in the evaluation set match for that given parameter, and -1 otherwise. We include maximal random effect slope structures, excluding correlations to help convergence, as per \citet{barr2013random}.

Our resulting regression model is reported in in \Cref{fig:reg2}, which we fit to obtain the reported $\beta$ coefficients, and corresponding p-values (Table \ref{tab:regression_results_exp2}).

\section{Frequencies}\label{sec:appendix-UD}

To calculate frequencies, we use the English-EWT Universal Dependencies dataset \citep{de2021universal, nivre-etal-2020-universal, silveira-etal-2014-gold}. 
It is sourced from the English Web Treebank, a corpus which totals 16,622 sentences scraped from the web. We parse the train, test, and dev \texttt{CoNLL-U} associated files searching for dependency relations denoting each of our given constructions. We do not differentiate between our two classes of embedded wh-questions, as the lexically defined constraint would have likely yielded a non-exhaustive extraction of all possible sentences. Instead we calculate a generic total for embedded wh-questions, and share this count among both of them. We present the final counts in Table \ref{tab:UD}.

\begin{table}[H]
    \centering
    \begin{tabular}{l r}
    \toprule
    \textbf{Construction Type}           & \textbf{Total Count} \\
    \midrule
    Restrictive Relative Clauses         & 504   \\ 
    Embedded Wh-Questions                & 308     \\ 
        Matrix Wh-Questions                  & 82                   \\  
    Clefts               & 20                   \\ 
    Pseudo-Cleft         & 6                    \\ 
    Topicalization      & 6                                  \\ 
    \midrule
    \textbf{Total Sentences} & \textbf{16622} \\ 
    \bottomrule
    \end{tabular}
    \caption{Construction Type Counts}
    \label{tab:UD}
\end{table}

\section{Experiment 1: Supplementary Information}\label{app:exp1-supp}

\begin{figure*}[p]
    \centering
    \includegraphics[width=.98\linewidth]{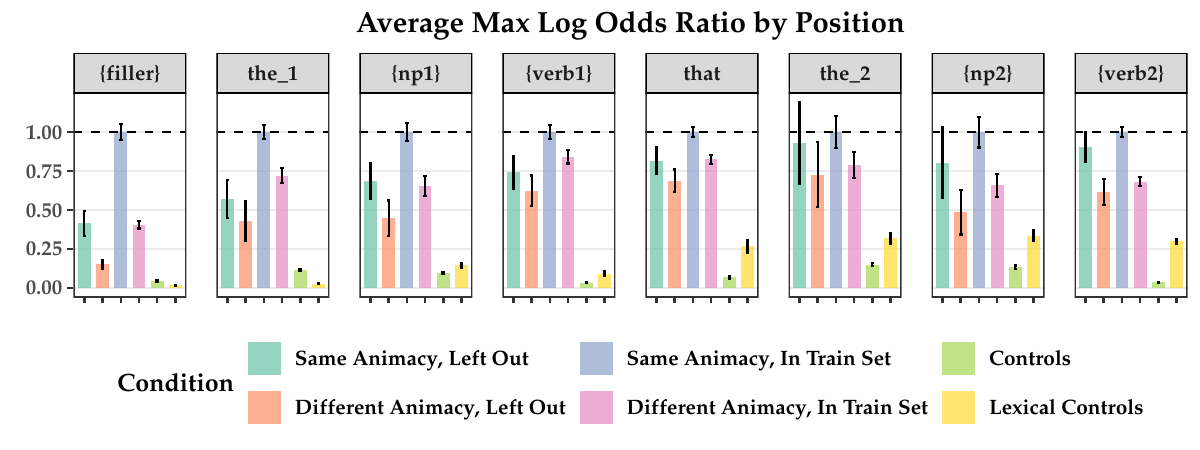}
    \caption{Multi-Clause Aggregation Values by Evaluation Group}
    \label{fig:loo-combined-embedded}
\end{figure*}

\begin{figure*}[p]
    \centering
    \includegraphics[width=.98\linewidth]{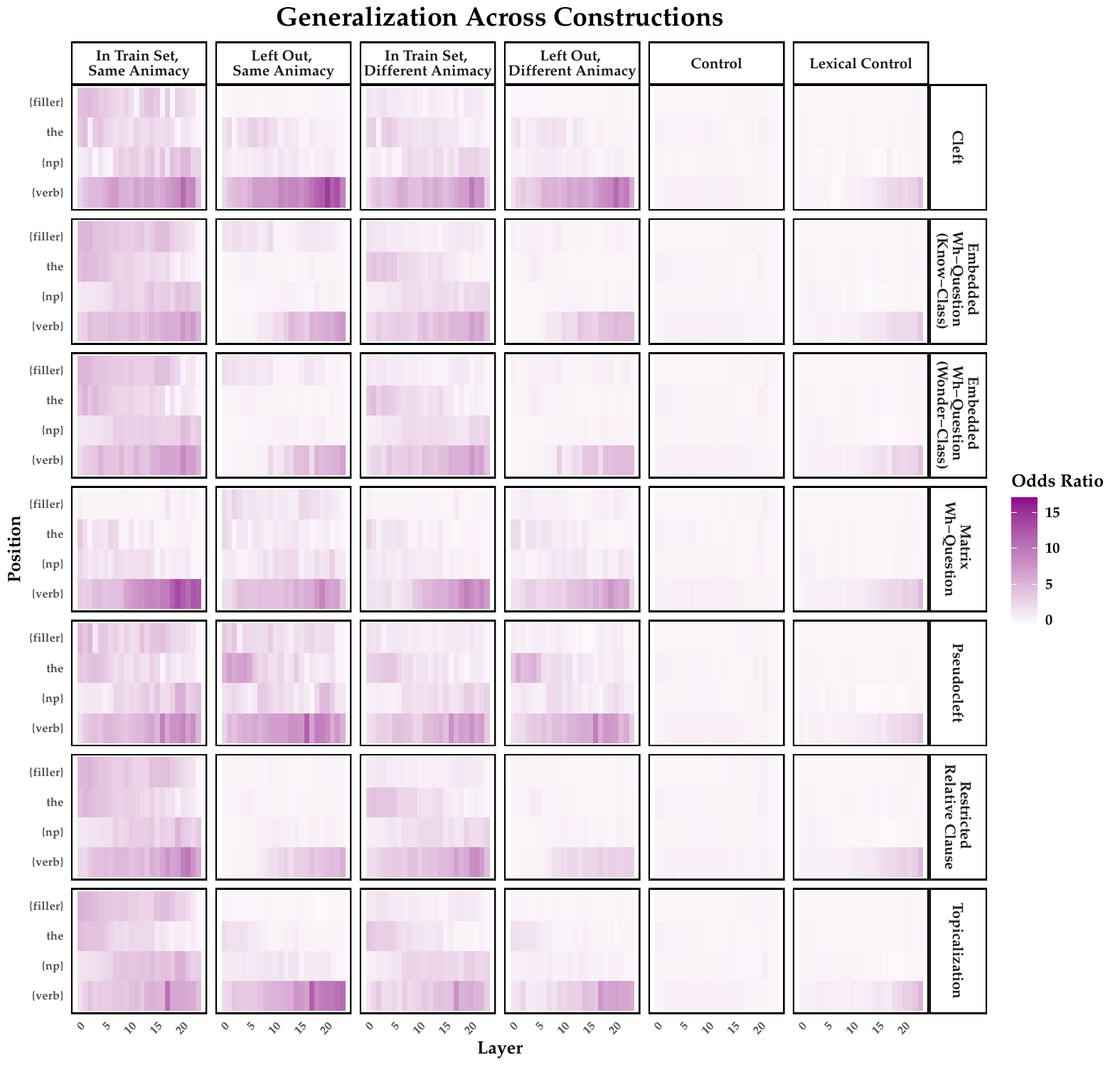}
    \caption{Single Clause \textsc{\textbf{odds}} at each position-layer pair for each construction. Averaged across animacy conditions.}
    \label{fig:mech_loo_control}
\end{figure*}

\begin{figure*}[ht]
    \vspace{0pt}
    \centering
    \includegraphics[width=\linewidth]{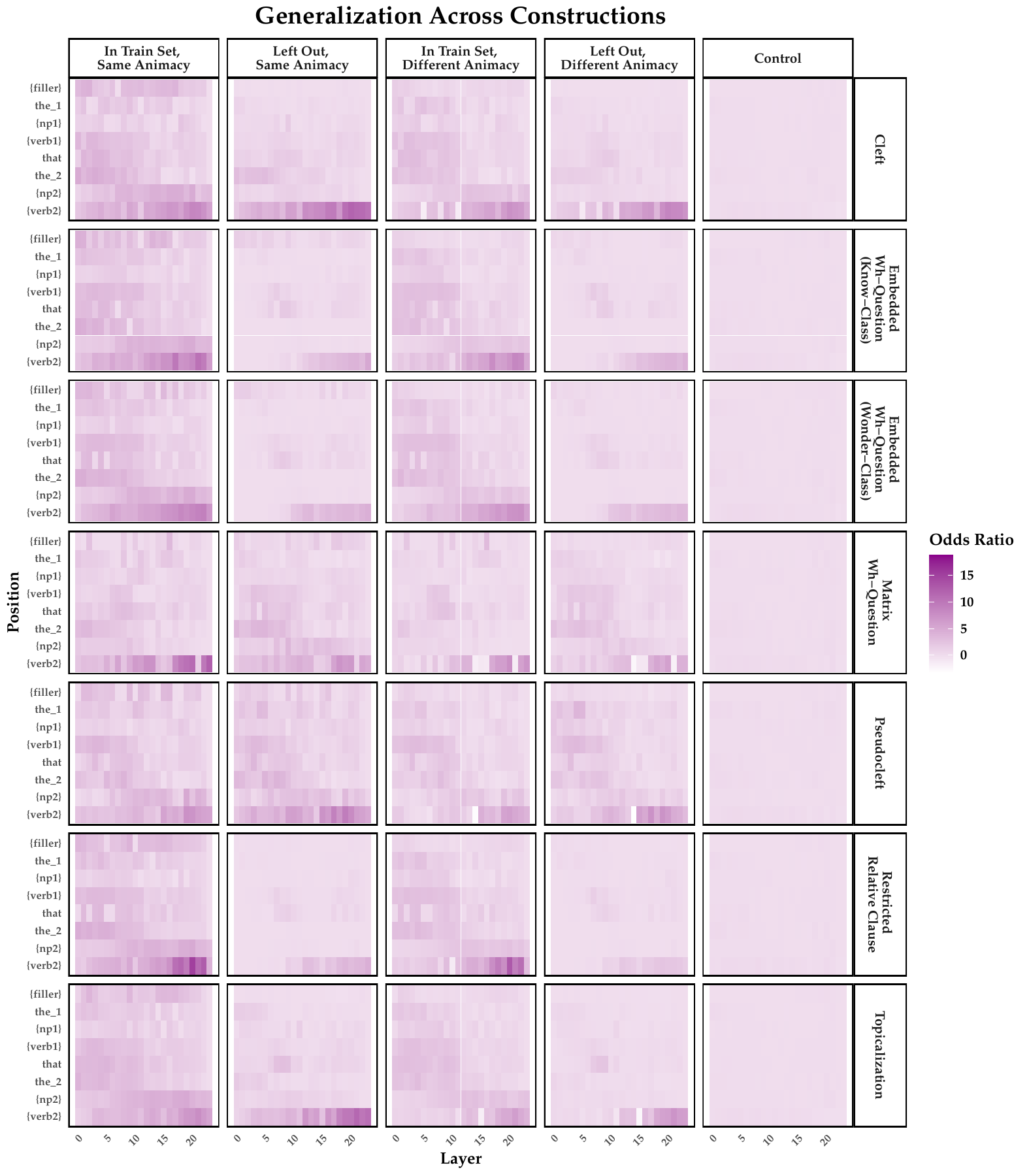}
    \caption{Multi-Clause \textsc{\textbf{odds}} at each position-layer pair for each construction. Averaged across animacy conditions.}
    \label{fig:mech_loo_control_embed}
\end{figure*}

A by-position aggregation figure for the multi-clause variant is in \Cref{fig:loo-combined-embedded}, complementing \Cref{fig:loo-combined}. An extended version of the mechanistic plots in \Cref{fig:mech_loo}, including controls, appears in \Cref{fig:mech_loo_control}, with a multi-clause counterpart shown in \Cref{fig:mech_loo_control_embed}.

\section{Experiment 2: Supplementary Information}\label{app:exp2-supp}

\begin{figure*}[!tp]
    \centering
    \includegraphics[width=.98\linewidth]{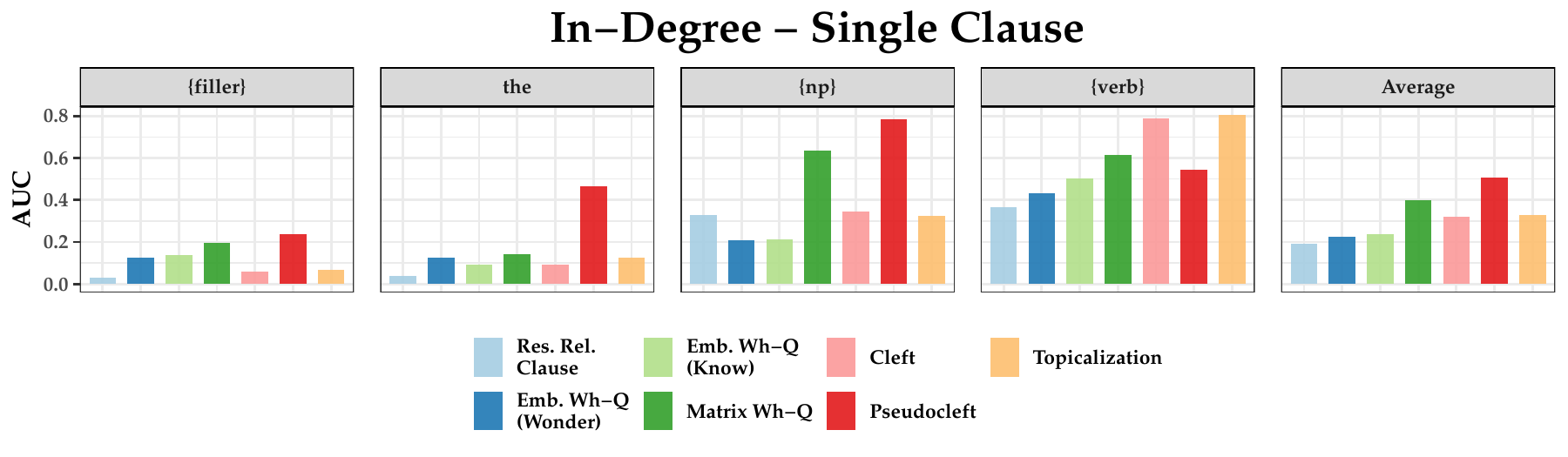}
    \caption{In-Degree AUC by position, with the final facet denoting the average across positions.}
    \label{fig:in_single}
\end{figure*}

\begin{figure*}[p]
    \centering
    \includegraphics[width=.98\linewidth]{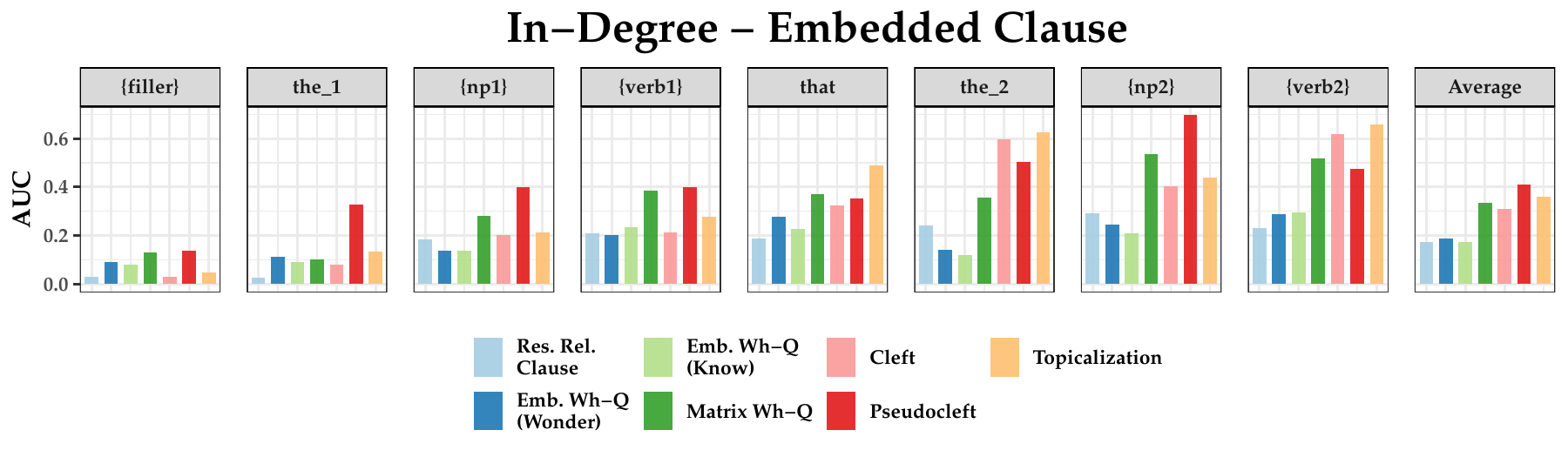}
    \caption{In-Degree AUC by position, with the final facet denoting the average across positions.}
    \label{fig:in_multi}
\end{figure*}

\begin{figure*}[p]
    \centering
    \includegraphics[width=.98\linewidth]{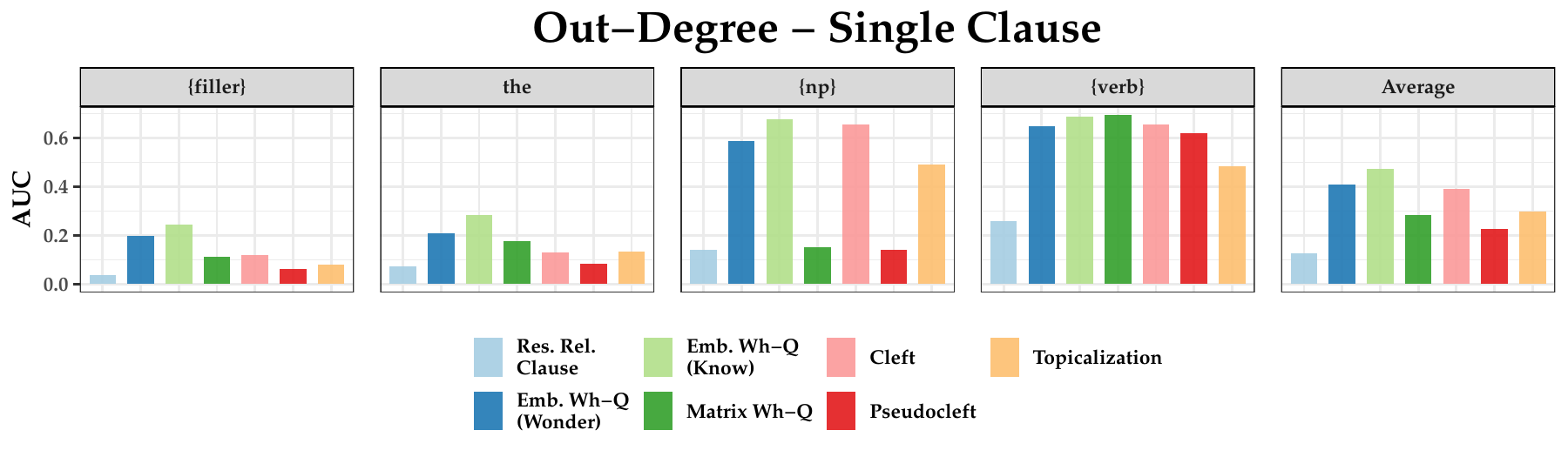}
    \caption{Out-Degree AUC by position, with the final facet denoting the average across positions.}
    \label{fig:out_single}
\end{figure*}

\begin{figure*}[p]
    \centering
    \includegraphics[width=.98\linewidth]{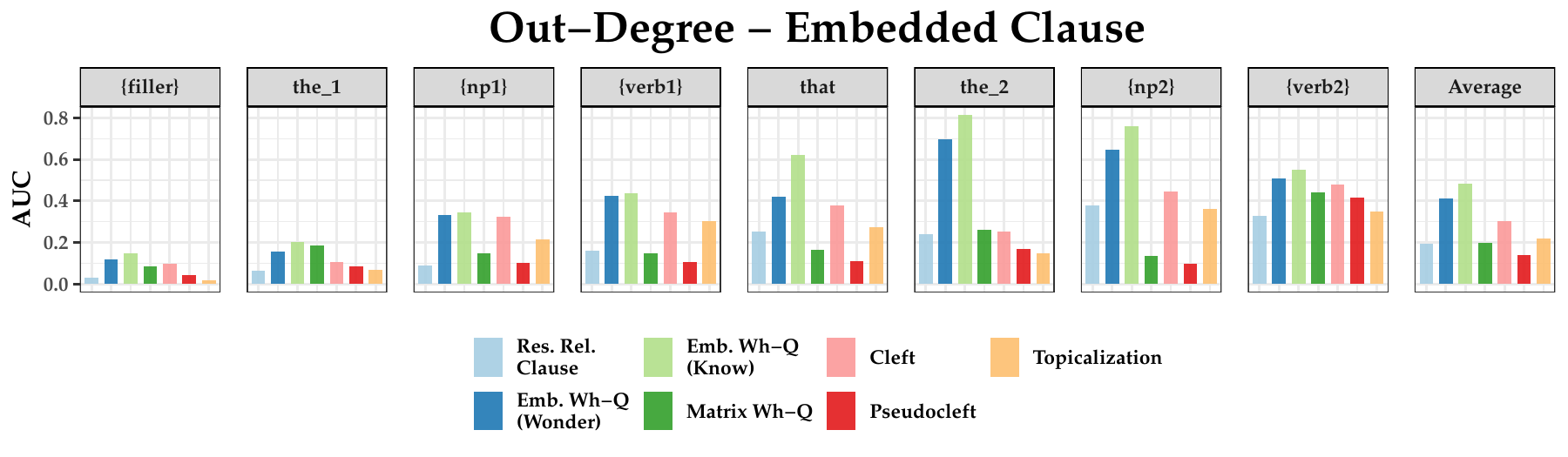}
    \caption{Out-Degree AUC by position, with the final facet denoting the average across positions.}
    \label{fig:out_multi}
\end{figure*}

We report raw bar charts for AUCs of in-degree and out-degree centrality across single- and multi-clause settings (\cref{fig:in_single,fig:in_multi,fig:out_single,fig:out_multi}). 


\section{Experiment 3: Supplementary Information}\label{app:exp3-supp}
\begin{figure*}[ht]
    \centering
    \includegraphics[width=.95\linewidth]{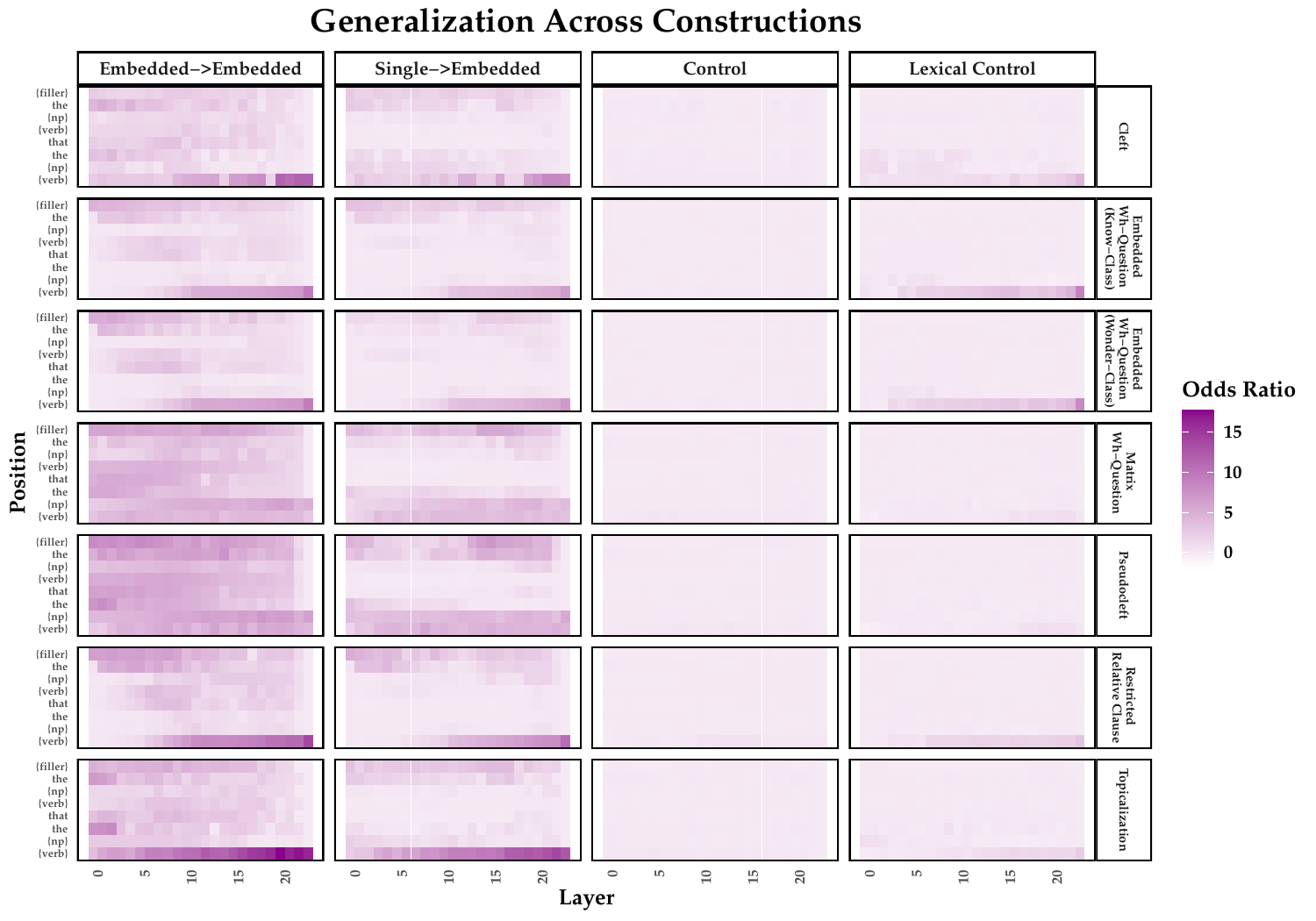}
    \caption{\textsc{\textbf{odds}} at each position-layer pair for each construction in the cross-clausal generalization experiment. Averaged across animacy conditions and items in a given group.}
    \label{fig:clausal}
\end{figure*}

We also provide mechanistic heatmaps for our cross-clausal generalization experiments. They can be found in \Cref{fig:clausal}.

\section{Replication with Other Model Sizes}\label{app:other-sizes}

We replicate these experiments with other model sizes, namely \texttt{pythia-2.8b} and \texttt{pythia-6.9b}. Below, we report these results. 
\newpage
\subsection{Experiment 1}
\begin{figure*}[htp]
    \centering
    \begin{subfigure}{1\linewidth}
        \centering
        \includegraphics[width=\linewidth]{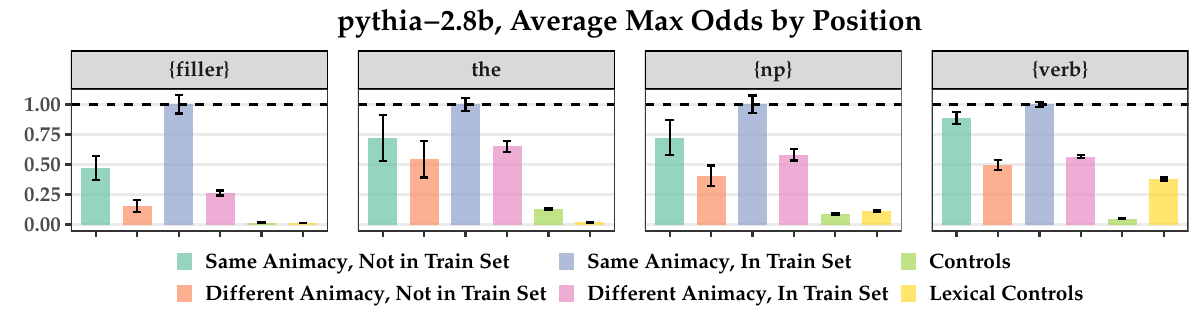}
        \caption{\texttt{pythia 2.8b} average normalized \textsc{\textbf{max odds}}.}
    \end{subfigure}
    
    \vspace{1em}
    
    \begin{subfigure}{1\linewidth}
        \centering
        \includegraphics[width=\linewidth]{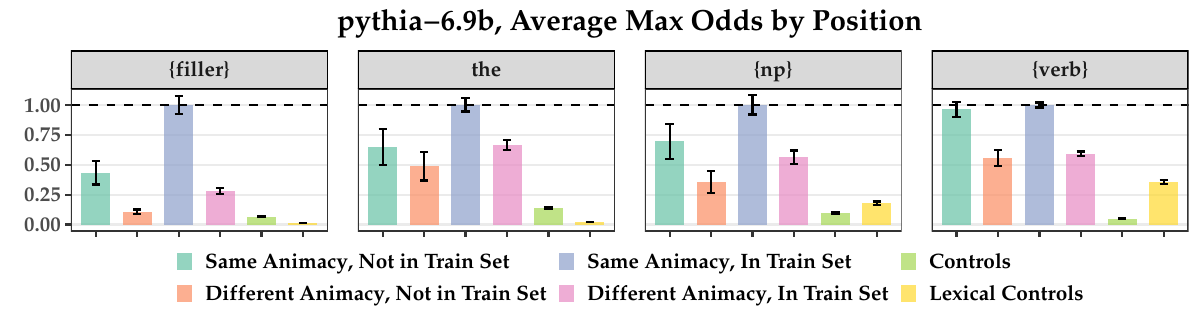}
        \caption{\texttt{pythia 6.9b} average normalized \textsc{\textbf{max odds}}.}
    \end{subfigure}
    
    \caption{\textbf{Top:} \texttt{pythia-2.8b} and \textbf{bottom:}  \texttt{pythia-6.9b} average normalized \textsc{\textbf{max odds}} across positions in the single-clause variants, $\pm 1$ standard error. Normalization fixes the ``Same Animacy, In Train Set'' condition at 1.00.}
    \label{fig:exp1_agg_single}
\end{figure*}

\begin{figure*}[htp]
    \centering
    \begin{subfigure}{1\linewidth}
        \centering
        \includegraphics[width=\linewidth]{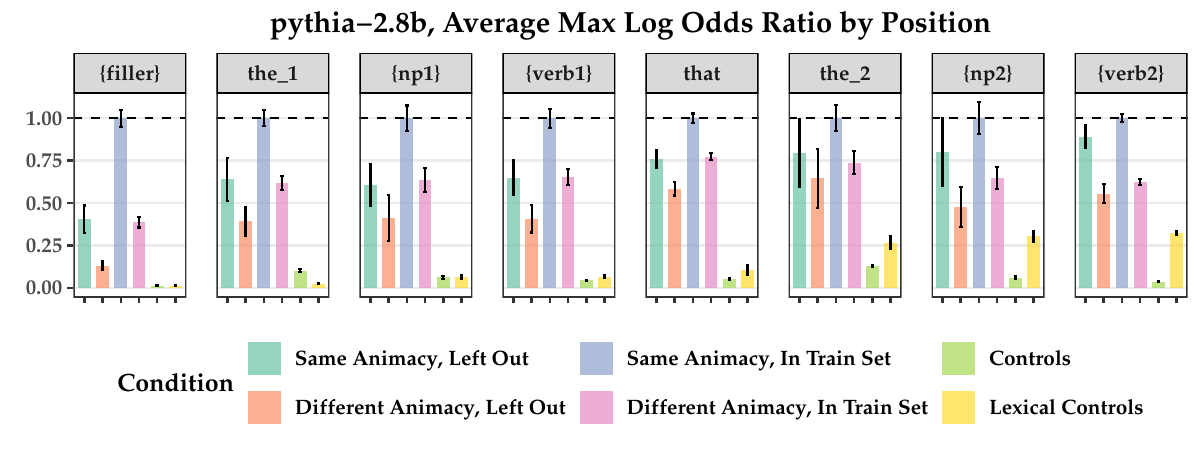}
        \caption{\texttt{pythia 2.8b}}
    \end{subfigure}
    
    \vspace{1em}
    
    \begin{subfigure}{1\linewidth}
        \centering
        \includegraphics[width=\linewidth]{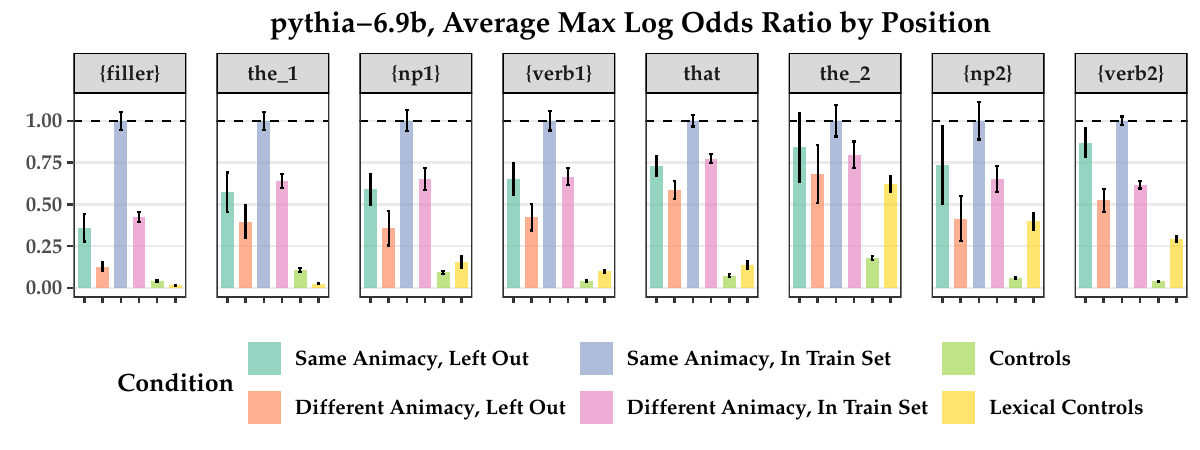}
        \caption{\texttt{pythia 6.9b}}
    \end{subfigure}
    
    \caption{\textbf{Top:} \texttt{pythia-2.8b} and \textbf{bottom:}  \texttt{pythia-6.9b} average normalized \textsc{\textbf{max odds}} across positions in the single-clause variants, $\pm 1$ standard error. Normalization fixes the ``Same Animacy, In Train Set'' condition at 1.00.}
    \label{fig:exp1_agg_embed}
\end{figure*}

We provide the aggregation figures across positions -- single (\Cref{fig:exp1_agg_single}) and multi-clause (\Cref{fig:exp1_agg_embed}) variants. We note that we find significant differences in the same positions as with the \texttt{pythia-1.4b} models. We provide regression results in \Cref{tab:regression_results_exp1_larger}.

\begin{table}[ht]
\centering
\resizebox{\columnwidth}{!}{%
\scriptsize
\begin{tabular}{lrrrr}
\toprule
\textbf{Term} & \textbf{$\beta$}$_{\textsc{filler}}$ &  \textbf{$\beta$}$_{\textsc{the}}$ &  \textbf{$\beta$}$_{\textsc{np}}$ &  \textbf{$\beta$}$_{\textsc{verb}}$ \\
\midrule
\multicolumn{3}{l}{\texttt{pythia-2.8b}} \\
(Intercept)                  & \textbf{1.95***} & \textbf{2.74***} & \textbf{1.83***} & \textbf{7.68***}  \\
\texttt{in\_train\_set}      & \textbf{0.67***} & \textbf{0.50***} & \textbf{0.37**} & \textbf{0.48*}  \\
\texttt{same\_animacy}       & \textbf{1.08***} & \textbf{0.51***} & \textbf{0.51***} & \textbf{2.18***}  \\
\texttt{in\_train\_set:same\_animacy} & \textbf{0.45**} & \textbf{0.19**} & 0.09 & 0.13  \\
\midrule
\multicolumn{3}{l}{\texttt{pythia-6.9b}} \\
(Intercept)                           & \textbf{1.78***} & \textbf{2.59***} & \textbf{1.48***} & \textbf{9.15***}  \\
\texttt{in\_train\_set}               & \textbf{0.76***} & \textbf{0.59***} & \textbf{0.36**} & 0.20 \\
\texttt{same\_animacy}                & \textbf{1.05***} & \textbf{0.47***} & \textbf{0.46***} & \textbf{2.45***}  \\
\texttt{in\_train\_set:same\_animacy} & \textbf{0.42**} & \textbf{0.18**} & 0.07 & 0.00  \\
\bottomrule
\end{tabular}
}
\caption{Experiment 1 Regression Results for \texttt{pythia-2.8b} and \texttt{pythia-6.9b}. * denotes $p<.05$, ** denotes $p<.01$, and *** denotes $p<.001$.}
\label{tab:regression_results_exp1_larger}
\end{table}
\vspace{-2em}
\subsection{Experiment 2}

\begin{figure*}[htp]
    \centering
    \begin{subfigure}[t]{0.48\textwidth}
        \centering
        \includegraphics[width=\linewidth]{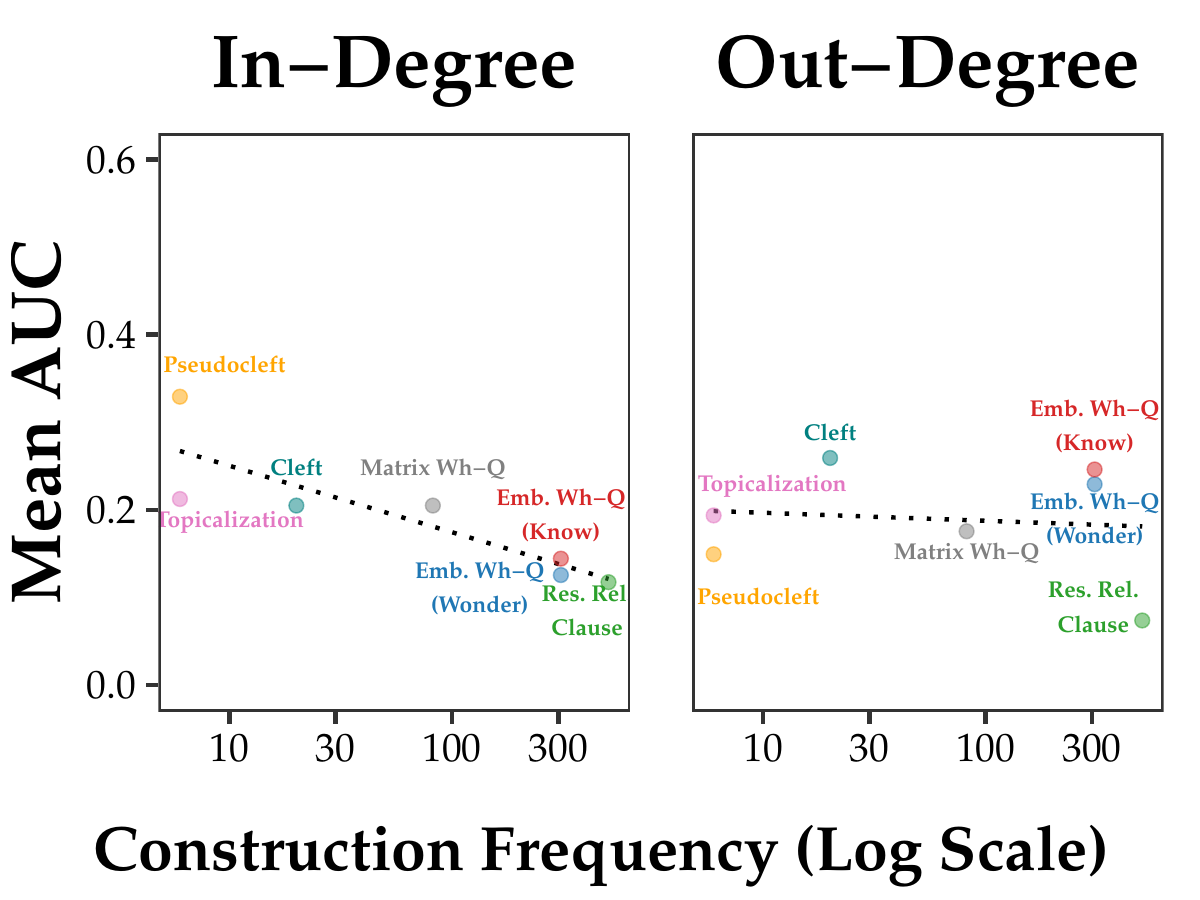}
        \caption{\texttt{pythia 2.8b}}
    \end{subfigure}
    \hfill
    \begin{subfigure}[t]{0.48\textwidth}
        \centering
        \includegraphics[width=\linewidth]{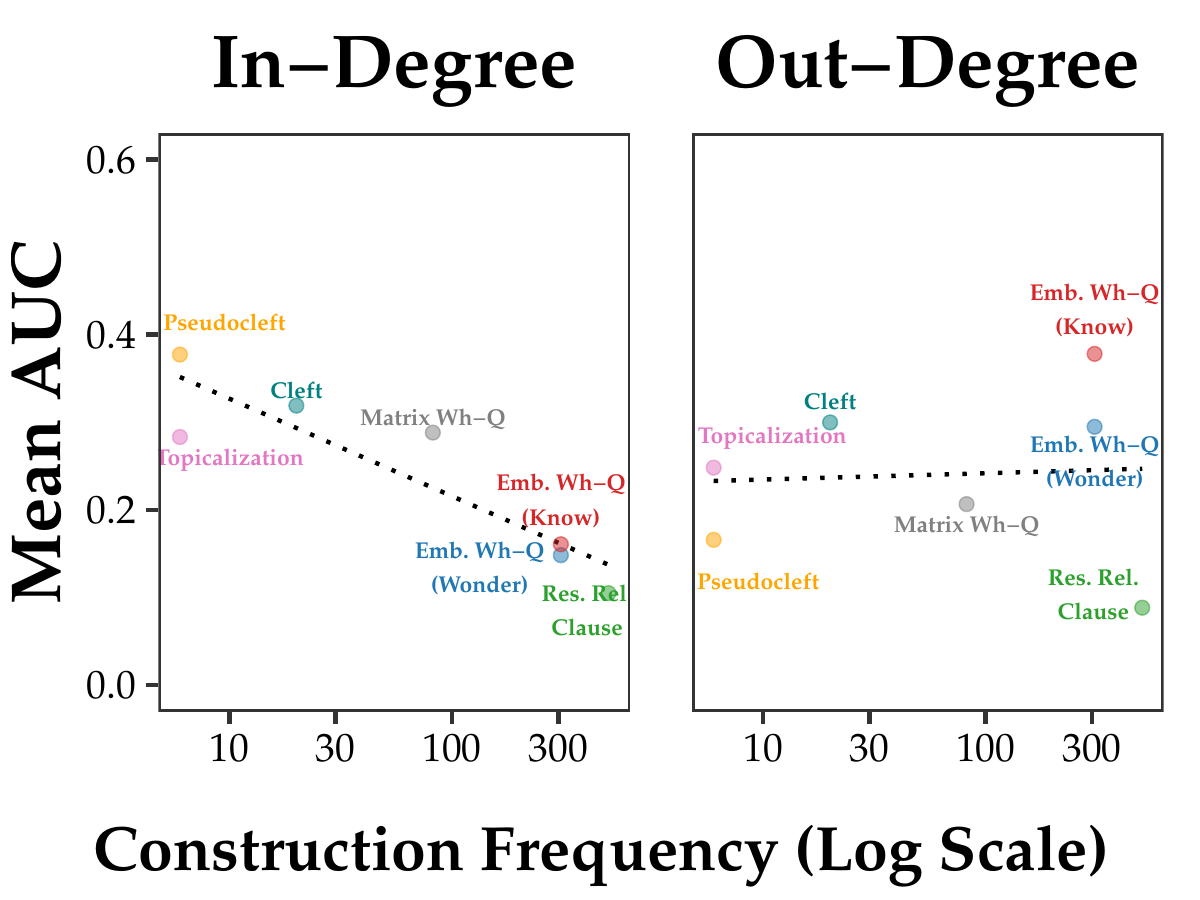} 
        \caption{\texttt{pythia 6.9b}}
    \end{subfigure}
    \caption{Average in-degree centrality AUC and out-degree centrality AUC plotted against construction frequency.}
    \label{fig:exp2_scatter}
\end{figure*}

\begin{figure*}[htp]
    \centering
    \begin{subfigure}[t]{0.48\textwidth}
        \centering
        \includegraphics[width=\linewidth]{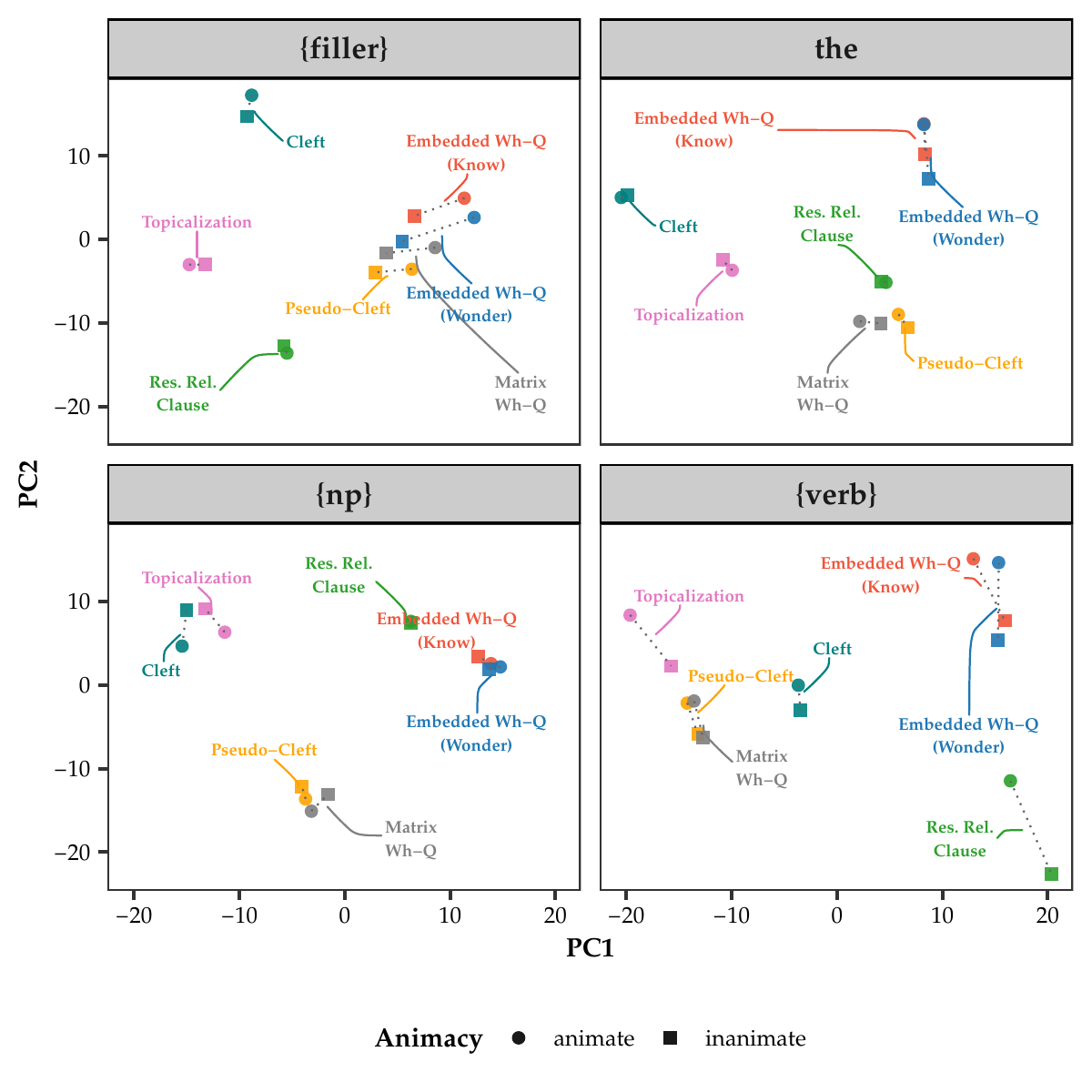}
        \caption{\texttt{pythia 2.8b}}
    \end{subfigure}
    \hfill
    \begin{subfigure}[t]{0.48\textwidth}
        \centering
        \includegraphics[width=\linewidth]{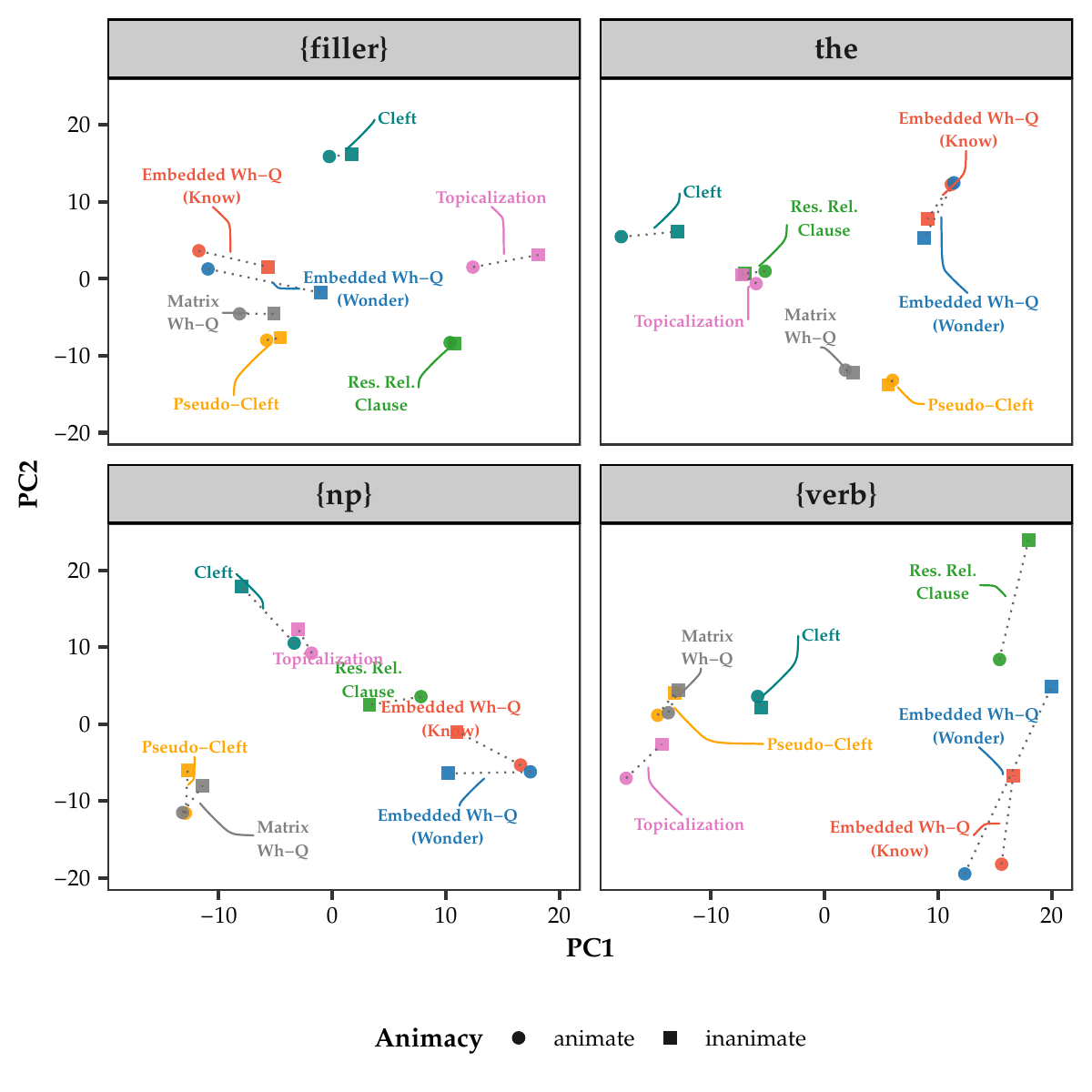} 
        \caption{\texttt{pythia 6.9b}}
    \end{subfigure}
    \caption{Constructions plotted along the top two principal components at each position in our single-clause variants.}
    \label{fig:pca_larger}
\end{figure*}

For experiment 2, we provide scatter plots in \Cref{fig:exp2_scatter}, regression results in Table \ref{tab:regression_results_exp2_larger}, and PCA plots in \Cref{fig:pca_larger}.

\vspace{-1em}
\begin{table}[H]
\centering
\resizebox{\columnwidth}{!}{%
\scriptsize
\begin{tabular}{lrrrr}
\toprule
\textbf{Term} & \textbf{$\beta$}$_{\textsc{filler}}$ &  \textbf{$\beta$}$_{\textsc{the}}$ &  \textbf{$\beta$}$_{\textsc{np}}$ &  \textbf{$\beta$}$_{\textsc{verb}}$ \\
\midrule
\multicolumn{3}{l}{\texttt{pythia-2.8b}} \\
(Intercept)                  & \textbf{1.05***} & \textbf{1.99***} & \textbf{1.20***} & \textbf{6.20***}  \\
\texttt{match\_filler\_class}      & \textbf{0.68***} & \textbf{1.16**} & 0.27 & \textbf{0.78**}  \\
\texttt{match\_inversion}         & \textbf{0.42**} & \textbf{0.46**} & \textbf{0.48***} & 0.29  \\
\texttt{match\_embedded\_under}   & \textbf{0.82***} & \textbf{1.03**} & \textbf{0.35***} & \textbf{1.95**}  \\
\texttt{match\_discourse\_fronted} & \textbf{0.30*} & 0.37 & \textbf{0.58**} & 0.32  \\
\midrule
\multicolumn{3}{l}{\texttt{pythia-6.9b}} \\
(Intercept)                  & \textbf{1.10***} & \textbf{1.84***} & \textbf{1.08***} & \textbf{7.61***}  \\
\texttt{match\_filler\_class}      & \textbf{0.62**} & \textbf{1.10**} & 0.31 & 0.28  \\
\texttt{match\_inversion}         & \textbf{0.36*} & \textbf{0.60**} & \textbf{0.48***} & 0.01  \\
\texttt{match\_embedded\_under}   & \textbf{0.82**} & \textbf{1.02**} & \textbf{0.53**} & \textbf{2.05**}  \\
\texttt{match\_discourse\_fronted} & \textbf{0.35*} & 0.31 & \textbf{0.39*} & 0.14  \\
\bottomrule
\end{tabular}
}
\caption{Experiment 2 Regression Results for \texttt{pythia-2.8b} and \texttt{pythia-6.9b}. * denotes $p<.05$, ** denotes $p<.01$, and *** denotes $p<.001$ .}
\label{tab:regression_results_exp2_larger}
\end{table}
\vspace{-1em}
\subsection{Experiment 3}

\begin{figure*}[tp]
    \centering
    \begin{subfigure}{\linewidth}
        \centering
        \includegraphics[width=\linewidth]{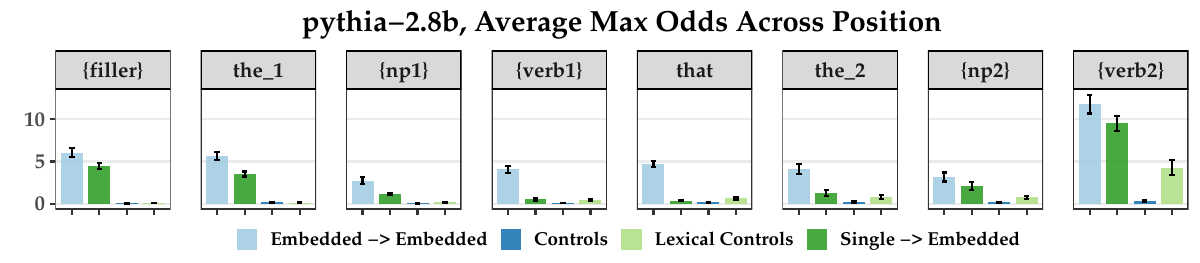}
        \caption{\texttt{pythia 2.8b}}
    \end{subfigure}
    
    \vspace{1em}
    
    \begin{subfigure}{\linewidth}
        \centering
        \includegraphics[width=\linewidth]{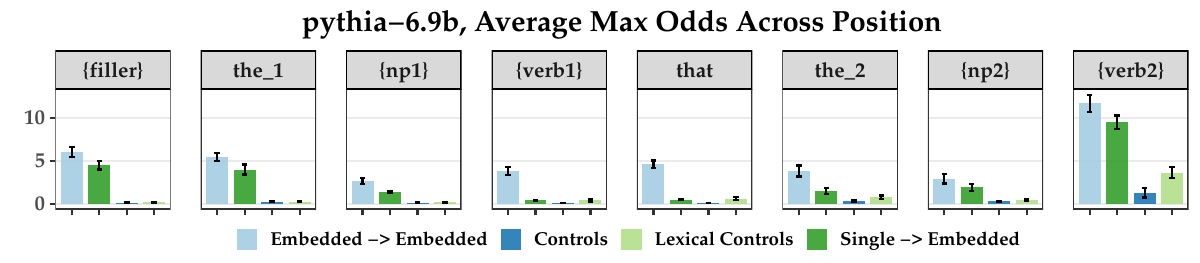}
        \caption{\texttt{pythia 6.9b}}
    \end{subfigure}
    
    \caption{\textsc{\textbf{max odds}} $\pm 1$ standard error, by position for interventions (1) trained and evaluated on multi-clause variants, (2) trained on single-clause variants and evaluated on multi-clause variants, and (3--4) controls. 
    Evaluations are performed on sentences matching training conditions (i.e. same construction and same animacy).}
    \label{fig:sd_other_sizes}
\end{figure*}

For experiment 3, we provide corollary figures to \Cref{fig:sd} in \Cref{fig:sd_other_sizes}.

\end{document}